\pdfoutput=1

\documentclass[format=sigconf,nonacm]{acmart}

\usepackage{microtype}
\usepackage{graphicx}
\usepackage{booktabs} 
\usepackage{balance}
\usepackage{enumerate}
\usepackage{tikz}
\usepackage{pgfplots}
\usepackage{pgfplotstable}
\usepackage{cleveref}
\usepackage{amsmath}

\usepackage{amssymb}
\usepackage{xcolor}
\usepackage{bm}
\usepackage{tabularx}
\usepackage{nicefrac}
\usepackage{siunitx}
\usepackage{subcaption}
\usepackage{colortbl}

\author{Rik Mulder}
\email{rik@rik.me}
\affiliation{University of Edinburgh}
\author{Valentin Radu}
\email{valentin.radu@ed.ac.uk}
\affiliation{University of Edinburgh}
\author{Christophe Dubach}
\email{christophe.dubach@ed.ac.uk}
\affiliation{University of Edinburgh}

\newcolumntype{L}{>{\raggedright\arraybackslash}X}
\pgfplotsset{compat=1.11,
	/pgfplots/ybar legend/.style={
		/pgfplots/legend image code/.code={%
			\draw[##1,/tikz/.cd,yshift=-0.25em]
			(0cm,0cm) rectangle (3pt,0.8em);},
	},
}

\pgfplotstableset{
	color cells/.style={
		col sep=comma,
		string type,
		postproc cell content/.code={%
			\pgfkeysalso{@cell content=\rule{0cm}{2.4ex}%
				\IfInteger{##1}{\edef\temp{\noexpand\cellcolor{red!##1}}\temp}{\edef\temp{\noexpand\cellcolor{red!0}}\temp}%
				##1}%
		}
	}
}
\definecolor{secondary}{rgb}{0.88671875,0.2890625,0.19921875}
\definecolor{primary}{rgb}{0.98828125,0.73046875,0.515625}

\definecolor{red1}{rgb}{0.400, 0.058, 0.082}
\definecolor{red2}{rgb}{0.800, 0.176, 0.149}
\definecolor{red3}{rgb}{0.984, 0.415, 0.290}
\definecolor{red4}{rgb}{0.988, 0.682, 0.568}
\definecolor{red5}{rgb}{0.996, 0.898, 0.850}

\definecolor{blue1}{rgb}{0.082, 0.058, 0.400}
\definecolor{blue2}{rgb}{0.149, 0.176, 0.800}
\definecolor{blue3}{rgb}{0.290, 0.415, 0.984}
\definecolor{blue4}{rgb}{0.568, 0.682, 0.988}
\definecolor{blue5}{rgb}{0.850, 0.898, 0.996}

\title{Optimising the Performance of Convolutional Neural Networks across Computing Systems using Transfer Learning}

\begin{document}

\begin{abstract}
The choice of convolutional routines (primitives) to implement neural networks has a tremendous impact on their inference performance (execution speed) on a given hardware platform.
To optimise a neural network by primitive selection, the optimal primitive is identified for each layer of the network. 
This process requires a lengthy profiling stage, iterating over all the available primitives for each layer configuration, to measure their execution time on the target platform. 
Because each primitive exploits the hardware in different ways, new profiling is needed to obtain the best performance when moving to another platform. 
In this work, we propose to replace this prohibitively expensive profiling stage with a machine learning based approach of performance modeling. 
Our approach speeds up the optimisation time drastically.
After training, our performance model can estimate the performance of convolutional primitives in any layer configuration.
The time to optimise the execution of large neural networks via primitive selection is reduced from hours to just seconds.
Our performance model is easily transferable to other target platforms. 
We demonstrate this by training a performance model on an Intel platform and performing transfer learning to AMD and ARM processor devices with minimal profiled samples.
\end{abstract}

\maketitle

\section{Introduction}\label{sec:intro}
Deep learning has triggered a machine learning renascence. 
This is now powering numerous applications, such as image classification, image segmentation, voice recognition, machine translation and others, which benefit billions of people.
However, these techniques are mostly designed for cloud computing on the server side.
This strategy is now increasingly scrutinized due to growing data privacy concerns and needing to operate on devices with sporadic Internet connectivity.

We are seeing a sustained adoption of deep learning in edge computing.
This is encouraged by increasingly more powerful hardware that can support complex applications locally, such as phones and smart home devices.
However, injurious use of powerful hardware is still limited in the breadth of workloads it can support and the rate of energy budget depletion on batter powered devices. 
Running deep neural networks at the edge on these devices requires compute adaptation to boost their efficiency.

Designing computationally efficient networks is one active research area in machine learning~\cite{squeezenet}. 
Already established large networks are often considered for deployment on smaller devices, by using quantization and model pruning~\cite{radu2019performance} adaptations.
But these methods reduce the original classification accuracy of those networks. 
An alternative approach to improve the execution of a neural network is to perform primitive selection at each convolutional layer in the network.
This retains its original inference accuracy.
These primitives represent alternative implementations of the convolutional operation, each producing an equivalent data processing result, but different in performance (execution speed). 

The primitive selection process starts with a large set of available implementations --- the primitives.
Convolutional primitives include im2col with matrix multiplication, direct convolution, winograd techniques and others.
These are characterized by the algorithm they use, data input and output organisations, system characteristics (vectorization, etc.).
The goal is to determine the best primitive and its data transformation for each layer configuration of the network, which produces the fastest execution on a given device.

However, the primitive selection method requires a lengthy profiling stage to measure the actual execution time of each primitive in the network configurations on the target platform.
This makes primitive selection difficult to apply in most applications as network layer configurations are quite varied. 
Consider developing at scale a machine learning-powered application on a smartphone.
To optimise the network configurations, profiling is required on all possible CPUs and GPUs available on the market.
This is unimaginable for a one-off calibration in the factory, so profiling is done when installing the application instead. 
Each additional application that requires the use of their own efficient neural network would go through the same profiling stage. 
In our experience, the profiling of all the primitives with the layers of very large neural networks (ResNet), can take up to hours.

Here, we ease the highly expensive profiling stage (often measured in hours~\cite{primitive}), by replacing it with a performance model. 
Two multi-layer fully-connected neural network architectures are evaluated as our performance model, which we show outperform simpler regression models. 
These are also ideal for transfer learning.
We show that once a performance model is trained on a given platform, very few additional sample points are required from the target platform (different architecture than original platform) to transfer the performance model. 

Intuitively, this eases the task of hardware vendors by building a performance model \emph{at the factory} only once for each platform. 
This performance model can be easily ported to other subsequent devices via transfer learning.
When an \emph{application} registers its neural network to run on the device, our performance model will be used to find the best configuration of primitives for the network. 
The optimisation of a convolutional neural network with our performance model takes just milliseconds, contrasting with to hour-long exhaustive profiling of primitives on the device for each new network. 

\clearpage

This paper makes the following contributions:
\begin{enumerate}
	\item 
	We design a performance estimation model that accurately predicts the execution time of convolutional primitives for any configuration of a convolutional layer.
	\item
	We show that using a performance model for primitive selection offers a significant speedup in neural network optimisation compared to profiling the device.
	\item
	The performance model are shown to be easily transferable across platforms (hardware architecture), with efficient estimations on AMD and ARM architectures after pre-training the model on an x86 architecture.
\end{enumerate}

The rest of this paper is organized as follows.
\Cref{sec:rw} describes the primitive selection process and the previous work on performance modeling of neural networks using machine learning. 
The methods for performance modeling and primitive selection used in this work are discussed in \cref{sec:ml}.
The experiments are described in \cref{sec:exps} and their results are discussed in \cref{sec:results}.
Finally, \cref{sec:conclusion} concludes this work and provides some future work directions.

\section{Background and Related Work}\label{sec:rw}
Multiple methods exist for reducing the computational requirements of neural networks.
Solutions include reducing the floating point precision during inference \cite{low_float}, neural architecture search \cite{pppnet,nemo,monas,mnasnet} and specialized neural network compilers aiming to generate highly efficient machine code \cite{latte}.
Primitive selection is a particularity attractive solution for its simplicity.
Any existing network can be optimized using primitive selection without any application knowledge.

\subsection{Primitive Selection}\label{sec:theory_prim_rw}
\begin{figure}[t]
\centering
\begin{tikzpicture}
	\def\blockwidth{2}
	\def\blockheight{0.75}
	\def\blockspacey{1}
	\def\blockspacex{0.2}

	\newcommand{\primitive}[4]{
		\draw[draw=black,thin] (#1,#2) rectangle ++(\blockwidth,\blockheight) node[pos=.5] {#4};
	}

	\newcommand{\plines}[2]{
		\draw[opacity=0.2,thin,->] (#1,#2) -- (-\blockwidth*0.5-\blockspacex,#2+\blockspacey);
		\draw[opacity=0.2,thin,->] (#1,#2) -- (\blockwidth*0.5,#2+\blockspacey);
		\draw[opacity=0.2,thin,->] (#1,#2) -- (\blockwidth*1.5+\blockspacex,#2+\blockspacey);
	}

	\newcommand{\layerblock}[3]{
		\primitive{0}{#1}{#2.2}{B}
		\primitive{-\blockwidth-\blockspacex}{#1}{#2.1}{A}
		\primitive{\blockwidth+\blockspacex}{#1}{#2.3}{C}
	
		\node[left] at (-\blockwidth-\blockspacex,#1+\blockheight/2) {#2};
		\node[right] at (\blockwidth*2+\blockspacex,#1+\blockheight/2) {#3};
	}

	\newcommand{\linies}[2]{
		\node[right] at (\blockwidth*2+\blockspacex,#1+\blockheight+\blockspacey/2) {#2};	
		\node[above right] at (-\blockwidth-\blockspacex-1,#1+\blockheight+\blockspacey/2) {\footnotesize data transf.};	
		\plines{-\blockwidth*0.5-\blockspacex}{#1+\blockheight}
		\plines{\blockwidth*0.5}{#1+\blockheight}
		\plines{\blockwidth*1.5+\blockspacex}{#1+\blockheight}
	}

	\newcommand{\inlines}[1]{
		\draw[opacity=0.2,thin,->] (\blockwidth*0.5,#1) -- (\blockwidth*0.5,#1+\blockspacey/3);
		\draw[opacity=0.2,thin,->] (\blockwidth*1.5+\blockspacex,#1) -- (\blockwidth*1.5+\blockspacex,#1+\blockspacey/3);
		\draw[opacity=0.2,thin,->] (-\blockwidth*0.5-\blockspacex,#1) -- (-\blockwidth*0.5-\blockspacex,#1+\blockspacey/3);
	}

	\inlines{-\blockspacey/3};
	\layerblock{0}{Conv1}{$\lambda^{N_1}$};
	\linies{0}{$\lambda^{E_1}$}
	\layerblock{\blockheight+\blockspacey}{Conv2}{$\lambda^{N_2}$};
	\linies{\blockheight+\blockspacey}{$\lambda^{E_2}$}
	\layerblock{\blockheight*2+\blockspacey*2}{Conv3}{$\lambda^{N_3}$};
	\inlines{\blockheight*3+\blockspacey*2};
	
	
	
	\draw[dashed] (-\blockspacex-\blockwidth-1,\blockheight+0.5*\blockspacey) -- (\blockspacex+\blockwidth*2,\blockheight+0.5*\blockspacey);
	\draw[dashed] (-\blockspacex-\blockwidth-1,\blockheight*2+1.5*\blockspacey) -- (\blockspacex+\blockwidth*2,\blockheight*2+1.5*\blockspacey);
	
	\draw[thick,->] (\blockwidth/2,\blockheight) -- (\blockwidth/2,\blockheight+\blockspacey);
	\draw[thick,->] (\blockwidth/2,\blockheight*2+\blockspacey) -- (\blockwidth*1.5+\blockspacex,\blockheight*2+\blockspacey*2);
	\draw[thick,->] (\blockwidth*0.5,-\blockspacey/3) -- (\blockwidth*0.5,0);
	\draw[thick,->] (\blockwidth*1.5+\blockspacex,\blockheight*3+\blockspacey*2) -- (\blockwidth*1.5+\blockspacex,\blockheight*3+\blockspacey*2+\blockspacey/3);
			
	\draw[draw=black,thick] (0,0) rectangle ++(\blockwidth,\blockheight);	
	\draw[draw=black,thick] (0,\blockheight+\blockspacey) rectangle ++(\blockwidth,\blockheight);	
	\draw[draw=black,thick] (\blockwidth+\blockspacex,\blockheight*2+\blockspacey*2) rectangle ++(\blockwidth,\blockheight);
\end{tikzpicture}
 	\caption{
 		Primitive selection illustrated for a three-layer convolutional neural network where each layer can be implemented with three primitives (A, B and C).
 		This gives 27 possible implementations for the overall network. 
 		The nodes and edges have associated costs $\lambda^N\in\mathbb{R}^3$ (the running time of the primitives) and $\lambda^E\in\mathbb{R}^{3\times 3}$ (the running time of the data layout transformations) respectively.
 		Primitive selection aims to assign a primitive to each layer such to minimize the total sum of node and edge costs.
 		In the figure, primitive B is selected for the first two layers and primitive C for the final layer.
 		This gives a total cost of $\lambda^{N_1}_2+\lambda^{E_1}_{2,2}+\lambda^{N_2}_2+\lambda^{E_2}_{2,3}+\lambda^{N_3}_3$.
 	}\label{fig:psel}
\end{figure}
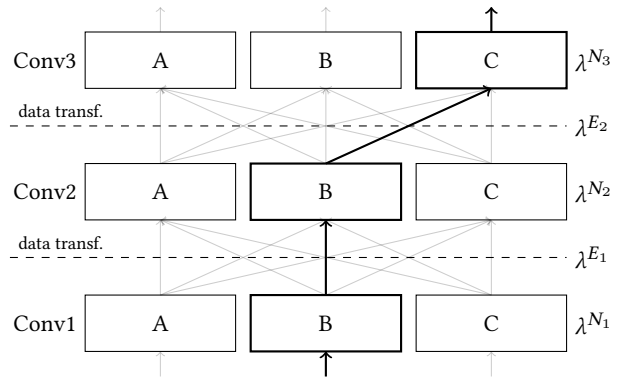

Primitive selection for a neural network means selecting which implementation -- or primitive -- to use for each of the layers in the network.
The assignment is made at layer level such that the entire running time of the network is optimized.
The running time of a primitive depends on the specific layer configuration and the hardware on which it is running.
This is further complicated by the fact that the various primitives can differ in the input and output data organisation. 
This captures the additional cost of `data-layout transformations' between layers.
Figure~\ref{fig:psel} represents schematically the process of primitive selection on a three-layer neural network. This chooses between primitive implementations $A$, $B$ and $C$ each with costs $\lambda^{E_{i}}_{A}$, $\lambda^{E_{i}}_{B}$ and $\lambda^{E_{i}}_{C}$ respectively, and data transformation between layers $\lambda^{N}$ when changing the primitive.

Previous work on primitive selection focuses on optimizing the execution time of convolutional neural networks by optimizing the primitives of each convolutional layer \cite{primitive,primitive_rl}.
Focusing only on the convolutional layers is justified by knowing that these layers incur more than 90\% of the execution time of modern neural networks~\cite{wang2019high}.
The problem is roughly solved in two stages.
First, all primitives are profiled on the target platform for all layers of the network (the profiling stage).
Depending on the network size and the target platform, this process can be very time consuming. 
Afterward, the execution time information is used to optimize the primitive selection (the optimization stage) for the entire network.

\paragraph{\textbf{The solver approach}}
In previous work~\cite{primitive}, the optimization stage is performed by an off the shelf Partitioned Boolean Quadratic Programming (PBQP) solver~\cite{pbqp}. This models the optimisation task as a graph traversal problem, in which primitives are represented by nodes with a cost and data transformations as edges with cost. 
For moderately sized networks such as Alexnet \cite{alexnet}, VGG-Net \cite{vgg} and Googlenet \cite{googlenet}, using the profiled data, a PBQP-solver can give optimal results in less than a second. 
Final solutions are shown to provide speed-ups of up to multiple times over common frameworks. Memory footprint has also been considered in the optimisation process with the Integer Linear Programming solver~\cite{wen2019poster,wen2020taso}.

\paragraph{\textbf{The reinforcement learning approach}}
This optimisation problem has also been explored by using reinforcement learning \cite{primitive_rl}.
This greatly reduces the search space by introducing heuristics about the cost of similar primitives or scaling cost on the size of the layer.
The limitation of this approach is that some primitives may behave very similar on one platform, but may have completely uncorrelated behaviour on other hardware or between layer sizes, which may lead to suboptimal solutions. 
By comparison with a stock optimisation library, the reinforcement learning approach showed a speed improvements of 1.4 $\times$.
However, there is no guarantee that the same heuristics would work on a different platform, so the additional cost of tuning the human-driven heuristics is still prohibitive.

Both works \cite{primitive,primitive_rl} rely on a lengthy profiling stage, which can make the method impractical to scale.
This work aims to remove the need for a profiling stage by introducing a performance model to predict the execution time of each primitive instead on a given hardware.

\begin{figure*}[t]
	\centering
	\begin{tikzpicture}[scale=0.7,every node/.style={scale=0.7}]
\draw[draw=black] (-8,2.4) rectangle ++(5,0.6);
\node[below right] at (-8,3) {\small\textbf{Conv}$_1$};
\node[below left] at (-3,3) {\small $k_1$, $c_1$, $im_1$, $f_1$, $s_1$};

\draw[draw=black] (-8,0.95) rectangle ++(5,0.6);
\node[below right] at (-8,1.55) {\small\textbf{Conv}$_2$};
\node[below left] at (-3,1.55) {\small $k_2$, $c_2$, $im_2$, $f_2$, $s_2$};

\draw[draw=black] (-8,-0.5) rectangle ++(5,0.6);
\node[below right] at (-8,0.1) {\small\textbf{Conv}$_p$};
\node[below left] at (-3,0.1) {\small $k_p$, $c_p$, $im_p$, $f_p$, $s_p$};

\draw[->] (-5.5,2.2) -- (-5.5,1.75);
\node at (-5.5,0.6) {\small$\vdots$};

\draw[dashed] (-6,-0.75) -- (-2.75,-0.75) -- (-2.75,3.25) -- (-6,3.25) -- (-6,-0.75);
\draw[->] (-2.5,1.25) -- node[below] {(i)} (-0.5,1.25);

\node at (-5.5,-1.5) {CNN to be optimized};
\node at (0,-1.5) {Layer configuration};
\node at (0,-1.85) {for every layer};
\node at (6,-1.5) {Runtimes predictions};
\node at (6,-1.85) {for all primtives for every layer};
\node at (12,-1.5) {Primitive assignment};
\node at (12,-1.85) {for every layer};

\foreach \name / \y in {c/0,k/0.5,im/1,f/1.5,s/2}
	\node[] at (0,2.25-\y) {\small $\vec{\bm{\name}}$};
\foreach \name / \y in {c/0,k/0.5,im/1,f/1.5,s/2}
	\draw[->] (0.3,2.25-\y) -- (0.8,2.25-\y);

\foreach \y / \k in {0/$\vec{\bm{R_1}}$,0.5/$\vec{\bm{R_2}}$,1/$\vec{\bm{R_3}}$,1.5/\vdots,2/$\vec{\bm{R_n}}$}
	\node[] at (6,2.25-\y) {\small\k};
\foreach \y / \k in {0/R_1,0.5/R_2,1/R_3,2/R_n}
	\draw[->] (5.20,2.25-\y) -- (5.7,2.25-\y);

\draw (1,0) -- (5,0) -- (5,2.5) -- (1,2.5) -- (1,0);
\node at (3,1.25) {\Large NN};
\node[above right] at (1,0) {(ii)};

\foreach \y / \k in {0/$\vec{\bm{R_1}}$,0.5/$\vec{\bm{R_2}}$,1/$\vec{\bm{R_3}}$,2/$\vec{\bm{R_n}}$}
\draw[->] (6.3,2.25-\y) -- (6.8,2.25-\y);

\foreach \y / \k in {0.25/$P_1$,0.75/$P_2$,1.25/\vdots,1.75/$P_p$}
\node[] at (12,2.25-\y) {\small\k};
\foreach \y / \k in {0.25/$P_1$,0.75/$P_2$,1.75/$P_p$}
\draw[->] (11.20,2.25-\y) -- (11.7,2.25-\y);

\draw (7,0) -- (11,0) -- (11,2.5) -- (7,2.5) -- (7,0);
\node at (9,1.25) {\Large PBQP};
\node[above right] at (7,0) {(iii)};

\draw[dashed] (11.5,0.15) -- (12.5,0.15) -- (12.5,2.35) -- (11.5,2.35) -- (11.5,0.15);
\draw[dashed,->] (12,2.65) -- (12,4) -- node[below] {(iv)} (-7,4) -- (-7,3.3); 

\end{tikzpicture}
	\caption{The primitive selection process when using a neural network based performance model. $p$ denotes the number of layers of the CNN. $n$ denotes the total number of primitives and data-layout transformations (DLT). The four steps are as follows. (i) the configurations of the layers (input number of channels $c_i$, input size $im_i$, kernels number of channels $k_i$, kernel size $f_i$ and stride $s_i$) are extracted. (ii) the configurations are given to the performance model which outputs, for every configuration, the predicted primitive and DLT running times $R_i$. (iii) the running times are given to a PBQP-solver which outputs the optimal primitive assignment for each layer $P_i$ for the network. (iv) the primitive assignment is used to optimize the running time of the original CNN. Note that the performance model is batched -- performing the computation for all layer configurations simultaneously -- making its inputs and outputs vectors of size $p$.}
	\label{fig:ml_ps}
\end{figure*}
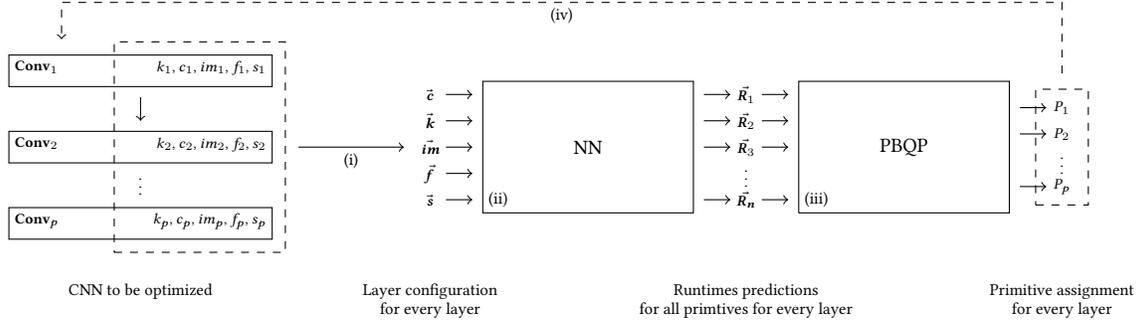

\subsection{Machine Learning for Running Time Prediction}\label{sec:rw_ml}
Previous work has shown that machine learning methods can predict the running time of algorithms via performance models~\cite{runtime_pred_alogs,benchmark_ml}.
Neural networks are especially suited for this task \cite{benchmark_ml}.
Some effort has been invested in predicting the execution time of convolutional networks also~\cite{cnn_power_runtime_pred,cnn_cost_prediction}.
Both linear regressions with polynomial features \cite{cnn_power_runtime_pred} and fully connected neural network \cite{cnn_cost_prediction} have been shown to accurately predict the running time of a convolutional network. 
These are trained and evaluated for a single default primitive and never considered for different primitive candidates. 

Transfer learning methods have also been shown to significantly reduce the required number of data points when creating a performance model for benchmark workloads to match the characteristics of a new platform \cite{benchmark_ml}. 
This shows that transfer learning works well for benchmark performance models because of the wide variation between benchmarks, but the task we explore here is harder because of the subtle performance difference across various primitives. 

No other work has approached deep neural network optimisation at the primitive level via performance modeling, which is the scope of our work here. In the following section we present our approach.

\section{CNN Performance Optimisation}\label{sec:ml}
As mentioned in \Cref{sec:rw}, the main drawback of primitive selection methods is the requirement of a lengthy profiling stage to gather the execution cost of each primitive and data transformation.
This work aims to replace the need for a profiling stage before the primitive selection with a performance model that can predict the execution time of primitives and data transformations. 
We adopt a neural network based performance model due to its characteristics to capture subtle differences in performance.
We perform this estimation with a model trained for the target device and pass the estimated costs to a solver for primitive selection for the convolutional neural network.

Given information about a convolutional layer (i.e. the input size, the stride, padding, kernel size, number of kernels) the performance model predicts the execution time for all primitives for this layer.
Repeating the process for all layers of the network, we generate the graph required by the PBQP-solver~\cite{pbqp} to optimise the primitive selection across the network for best performance.

The entire primitive selection process is shown in Figure \ref{fig:ml_ps}:
\begin{enumerate}[i]
	\item Given a convolutional layer, the parameters of its configuration are extracted.
	\item The layer configuration information is used by the performance model to estimate the execution time of each primitive and each data-layout transformation.
	\item The execution time predictions are passed to the PBQP-solver to produce the primitive selection across the network.
	\item The primitive assignment is used to produce the final implementation for the optimised convolutional neural network.
\end{enumerate}

\subsection{Convolutional Layer Optimisation}\label{sec:theory_conv}
There are various ways in which a two-dimensional convolution can be performed. We consider the layer parameters: width $w$, height $h$ and number of channels $c$ of the input; kernel size $f$, stride $s$ and number of kernels $k$. For simplicity, we assumed that the input is square, which is common in most neural networks. The input size is notated as $\mbox{im}$ ($\mbox{im}=w=h$).

A significant number of highly efficient implementations of the convolutional operation have been proposed over the year, each with their drawbacks and benefits.
Some primitives may be restricted to just some data shapes, as $c\times h\times w$ whereas others may assume input images in a different format, $h\times w\times c$.
The output shape also varies between primitives, which is not necessarily the same as the input shape.
The major family of implementations (each which potentially multiple implementations) considered in this work are \textit{direct-sum2d}, \textit{im2}, \textit{kn2}, \textit{winograd}, \textit{conv-1x1} and \textit{mec}.
A list of all the primitives we used is presented in the appendix (\cref{tab:ap_prims}).

The \textbf{direct-sum2d} family computes the convolution through six nested for-loops, three to cover all output values ($w$, $h$ and $k$) and three for all input values that influence the specific output value ($f_w$, $f_h$ and $c$).
Given the lack of optimized routines, the direct-sum2d is often one of the slowest primitives with general compilation.
The \textbf{im2} (or image to) family, on the other hand, performs a convolution by reshaping the input data and kernels such that the convolution is performed through a single matrix multiplication.
This can greatly speed up the running time of the convolution \cite{im2}.
As there is substantial data replication, this can be quite memory intensive, but we are not concerned about the memory space here, although data access is captured in the primitive execution cost.
The \textbf{kn2} (or kernel to) primitives aim to reduce this memory inefficiency by not performing the entire convolution as a single matrix multiplication. 
Instead, the entire convolution is broken down into the sum of multiple matrix multiplications \cite{im2var}. 
The downside for the kn2 family is that it is not efficient for larger strides convolutions.

The \textbf{winograd} primitives perform the unstrided convolution with the least amount of multiplications possible \cite{quickwino}.
Although this can offer significant speed-ups, it is difficult to anticipate when this is the case without profiling.
We have winograd primitives specialized for different kernel sizes: $3\times 3$ and $5\times 5$.
For kernels of size $1\times 1$, the entire convolution is essentially a single element-wise matrix multiplication.
The \textbf{conv-1x1} family implements them as such.
Finally, the \textbf{mec} (or memory-efficient convolution) primitives are two primitives that optimise memory requirements of a convolutional layer~\cite{im2var}. This is most often in detriment of execution time, but also occasionally on-pair with the other performance driven primitives.

\subsection{Profiler Dataset}\label{sec:ml_ds} 
\subsubsection{Convolutional Primitive Execution Time}
To train a machine learning model to predict the execution time of primitives, we need a dataset of some data points representing the execution of primitives on a target device.
We collect these data points by profiling many convolutional layer configurations.
Let the execution time measured for primitive $i$ be $R_i$, then our dataset is constructed as follows:
\[
\left(k, c, im, s, f\right)\rightarrow\left(R_1, R_2,\dots,R_N\right)
\]
where left side is the layer configuration ($k$, $c$, $im$, $f$, $s$) and right side the measured times for $N$ primitives profiled in hardware.
Not all primitives work for every configuration (e.g. a primitive may require a specific kernel size), hence some $R_i$ can be undefined.

We identify the common ranges of parameters found in a wide pool of networks, as presented in Table \ref{tab:params}. We consider input images of up to 299 by 299 pixels -- the largest input size used for ImageNet \cite{imagenet}.
Over all these possible configurations, the total space size is roughly 20 billion different variations.
Some configurations may not occur in modern convolutional networks. 
For example, the image size tends to reduced further in the network where the number of convolutions are higher. 
But we avoid to constrain the search space abruptly with arbitrary heuristics, leaving the search space largely available for a more general solution.
The specific data points are selected as follows. 
First, a set of $c$, $k$ and $im$ triplets are collected as they occur in a large variety of common architectures (\cref{tab:ap_archs}).
This results in 475 unique triplets.
Then, each of these triplets is combined with all combinations for the tuples $f$ and $s$ (as listed in \cref{tab:params}) and impossible values (e.g. $f>im$) are filtered out.
The number of data points collected can be seen in \cref{tab:exp_ds_size}.

\begin{table}
	\centering
	\caption{Common parameter values for convolutional layers considering popular CNNs applied to image net. $im$ denotes the width and height of a square input.}\label{tab:params}
	\begin{tabular}{@{}lll@{}}
		\toprule
		Parameter & Meaning & Common Range \\\midrule
		$\bm{k}$ & \#kernels & 1 to 2048 \\
		$\bm{c}$ & \#channels & 1 to 2048 \\
		$\bm{im}$ & image size & 7 to 299 \\
		$\bm{s}$ & stride & 1, 2 or 4 \\
		$\bm{f}$ & kernel size & 1 to 11 (odd)\\
		\bottomrule
	\end{tabular}
\end{table}

For our list of primitives, we adopt the primitives highlighted in other primitive selection works~\cite{primitive}. 
This includes a wide range of primitives across all the major algorithm families discussed in Section \ref{sec:theory_conv}.
We ignored `fast Fourier transform' primitives as our initial experiments showed them to be the slowest for almost all layer configurations.
A full list of considered primitives can be found in the appendix (\cref{tab:ap_prims}).

\begin{table}
	\centering
	\caption{The number of datapoints present in the primitive running time dataset for each primitive. A mapping to the exact primitive names can be found in \cref{tab:ap_prims}.}\label{tab:exp_ds_size}
	\begin{tabular}{@{}ll@{}}
		\toprule
		Primitives & \# data points \\
		\midrule
		direct, mec, im2 (a-d, m-p) & 4665 \\
		kn2, im2 (e-l, r-t) & 1974 \\
		wino3, conv-1x1 & 419 \\
		wino5 & 417 \\
		\bottomrule
	\end{tabular}
\end{table}

\subsubsection{Data Layout Transformation Cost}
Convolutional primitives have different requirements for the shape of input data.
If one primitive outputs the data in a different format than what the following layer requires, this data needs to be transformed to the suitable format.
This adds an additional cost to using those two consecutive primitives.
As such, a solver that optimizes the primitive selection needs the execution times for data-layout transformations to determine the optimal selection across the network.
This set of transformation costs can be profiled on the hardware for any shape and size of data passed between layers.
This cost depends only on the data size ($c$ and $im$) and data-layout, with fewer candidates than for the primitive profiling.
These profiler measurements are represented in our dataset as entries of the form:
\[
\left(c, im\right)\rightarrow\left(R_{i,1}, R_{i,2},\dots,R_{i,K}\right)
\]
where $R_{i,j}$ is the execution time to perform a transformation from format $i$, with $i \in [1,K]$, to format $j$, with $j \in [1,K]$, and $K$ is the total number of data-layout transformations.
For our set of primitives, there are three different data layouts: $c\times im\times im$, $im\times c\times im$ and $im\times im\times c$.
This results in nine data layout transformations to profile, which includes the identical transformation (to self) with cost zero.
All these measurements are profiled on a target hardware to form our data layout transformations dataset.

\subsection{Performance Modeling}\label{sec:ml_ml}
\paragraph{Performance Metrics}
We assess the quality of performance estimation using the median relative absolute error (MdRAE).
This gives a measure too which extend a predicted performance time is expected to be off.
The relative absolute error is defined as follows:
\[
\frac{\left|\hat{y}-y\right|}{y}
\]
where $\hat{y}$ represents the prediction and $y$ the actual measured value.    

\begin{figure}[t]
	\centering
	\begin{tikzpicture}[scale=0.8,every node/.style={scale=0.8}]

\node at (0,-0.3) {Take as input the};
\node at (0,-0.65) {shape of the convolutional};
\node at (0,-1) {layer to be optimised};

\node at (6.5,-1) {Estimated running time};
\node at (6.5,-1.35) {for each primitive separately};

\draw (1.5,-1) -- (4.5,-1) -- (4.5,-3.5) -- (1.5,-3.5) -- (1.5,-1);
\node at (3,-2.25) {\Large NN1$_{1}$};

\foreach \name / \y in {c/0,k/0.4,im/0.8,f/1.2,s/1.6}
	\node[] at (0.5,-1.5-\y) {\small $\vec{\bm{\name}}$};
\foreach \name / \y in {c/0,k/0.4,im/0.8,f/1.2,s/1.6}
	\draw[->] (0.8,-1.5-\y) -- (1.3,-1.5-\y);

\draw[->] (4.70, -2.25) -- (5.2,-2.25);
\node at (5.5,-2.25) {$\vec{\bm{R_1}}$};

\node at (3,-3.75) {\vdots};

\draw (1.5,-4.25) -- (4.5,-4.25) -- (4.5,-6.75) -- (1.5,-6.75) -- (1.5,-4.25);
\node at (3,-5.5) {\Large NN1$_{n}$};

\foreach \name / \y in {c/0,k/0.4,im/0.8,f/1.2,s/1.6}
	\node[] at (0.5,-4.75-\y) {\small $\vec{\bm{\name}}$};
\foreach \name / \y in {c/0,k/0.4,im/0.8,f/1.2,s/1.6}
	\draw[->] (0.8,-4.75-\y) -- (1.3,-4.75-\y);

\draw[->] (4.70, -5.5) -- (5.2,-5.5);
\node at (5.5,-5.5) {$\vec{\bm{R_n}}$};

\draw[dashed] (-1.5,-7.25) -- (8,-7.25);

\node at (6,-7.6) {Estimated running time};
\node at (6,-7.9) {for all primitives};

\foreach \name / \y in {c/0,k/0.5,im/1,f/1.5,s/2}
	\node[] at (0,-8.5-\y) {\small $\vec{\bm{\name}}$};
\foreach \name / \y in {c/0,k/0.5,im/1,f/1.5,s/2}
	\draw[->] (0.3,-8.5-\y) -- (0.8,-8.5-\y);

\foreach \y / \k in {0/$\vec{\bm{R_1}}$,0.5/$\vec{\bm{R_2}}$,1/$\vec{\bm{R_3}}$,1.5/\vdots,2/$\vec{\bm{R_n}}$}
	\node[] at (6,-8.5-\y) {\small\k};
\foreach \y / \k in {0/R_1,0.5/R_2,1/R_3,2/R_n}
	\draw[->] (5.20,-8.5-\y) -- (5.7,-8.5-\y);

\draw (1,-8.25) -- (5,-8.25) -- (5,-10.75) -- (1,-10.75) -- (1,-8.25);
\node at (3,-9.5) {\Large NN2};

\foreach \y / \k in {0/$\vec{\bm{R_1}}$,0.5/$\vec{\bm{R_2}}$,1/$\vec{\bm{R_3}}$,2/$\vec{\bm{R_n}}$};

\end{tikzpicture}
	\caption{We explore two performance models in the form of two multi-layer neural networks. These take as input the shape of the convolutional layer that needs to be optimised; (i) NN1 -- estimates the performance of each primitive separately; (ii) NN2 -- estimates the running time of all primitives at once. An ensemble of NN1 models produces the vector equivalent to NN2's output.}
	\label{fig:nn_i_ii}
\end{figure}
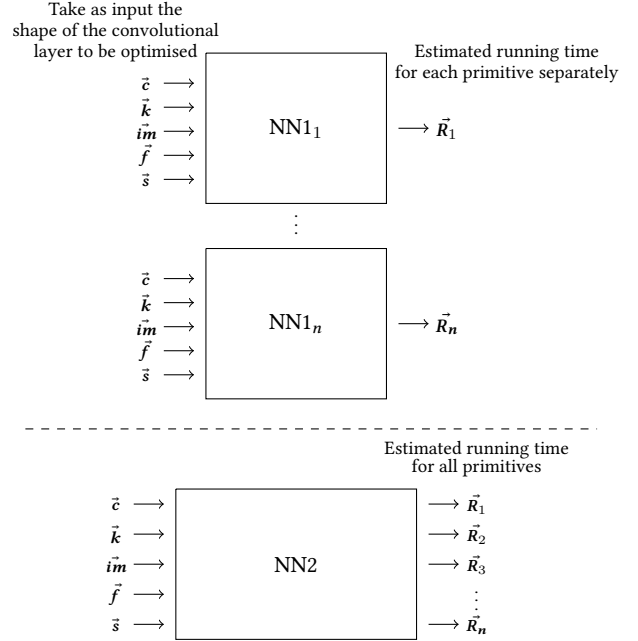

\paragraph*{Performance Model Architectures}
Given the non-linearity of most primitive families, a multi-layer fully-connected neural network is our choice for a good performance model.
We use two approaches to construct the neural network architecture (Figure~\ref{fig:nn_i_ii}).
The first (NN1), using a separate model for each primitive and for each data transformation.
In Figure~\ref{fig:nn_i_ii}, these are indicated as $NN1_{i}$, where $i$ is the order of primitives.
In the second (NN2), we use a single model to estimate the performance of all primitives. 
Both of these models take as input the parameters defining the shape of a convolutinal layer, as discussed in previous sections.
Step ii) of Figure \ref{fig:ml_ps} schematically represents our second approach, estimating all costs at once.
We emphasise NN2 because having a single model to estimate all performance costs makes deployment and logistics much simpler.
We also have to consider data-layout transformations, so a similar network is trained to predict the costs of these data transformations.
Using NN2 also reduces the training time as many similarities exist between primitives, which are automatically determined during training, thus reducing the amount of data required for training.

\paragraph*{Loss Function}
Predicting performance is a regression task. As such, we use the mean squared error (MSE) loss as the loss function for network output.
However, some execution times may be undefined, for instance when a primitive cannot be applied to a particular shape of a convolutional layer.
When training NN1 for each primitive, we filter out the undefined conditions. 
When training with NN2, for estimations of performance across all primitives, training sample points can have both actual values and undefined labels.
We make sure these have no effect on the training by masking the values and gradients of those primitives with undefined values in the forward pass and in the back-propagation stage respectively.
This causes their squared error to equate to zero, resulting in no influence on the training process.

\paragraph*{Data Point Normalization}
The performance times can be substantially wide in magnitude.
With our loss function, MSE, favorable estimations are observed in the larger values.
To address this issue, we adopt the log expression of performance values when training the network, to scale favorably both large and small values. 
Finally, to improve the training of the neural network, all input and output data were standardized (i.e. transformed to have zero mean and a standard deviation of one).
Performing all these steps would result in the following transformation for a given variable $x$:

\[
\tilde{x}=\frac{z-\bar{z}}{\mbox{std}\left(z\right)},\mbox{ where } z=\log(x)
\]

where $\tilde{x}$ denotes the normalized result and $\bar{z}$ the mean of $z$.

\section{Experiments}\label{sec:exps}
This section introduces the experiment setup for performance data collection. This is followed by preliminary observations on this data and how a performance model is calibrated. We also design experiments to validate that primitive selection can operate on predicted performance values and that a trained model can be transferred to use in other hardware.   

\subsection{Primitive Performance Profiling}
\subsubsection{Experimental Environment}\label{sec:exp_ds_env}
We run our data collection on three machines profiling the execution of all primitives. An Intel Core i9-9900K @ 5.0GHz, an AMD A10-7850K @ 3.7GHz and an ARM Cortex-A73 (rev2) @ 2.36GHz.
We use the primitive implementations from the open software triNNity-benchmarks\footnote{\url{https://bitbucket.org/STG-TCD/trinnity-benchmarks/}} (commit 8b0e642).
This is executed from a Docker container.
The benchmark is executed natively on the Intel and AMD machines.
For the ARM system, primitives are compiled with cross-compilation.

Primitives are profiled 25 times each to get the median value for reliability. 
In profiling, the layer configuration is run on a layer input with values drawn from a normal distribution.
This is to simulate any possible image input to the network, although the execution time is not much affected by input image.

\subsubsection{Preliminary Observations}
In either of the three datasets, no single primitive is routinely the quickest, neither is one family of primitives dominant in performance across all the layer configurations we tested with.
Even when fixing the value for $s$ (the stride) and $f$ (the kernel size), the quickest primitives are spread across all families of primitives. 
Overall, the fastest primitives are spread across many of the tested primitives, as discussed in Section \ref{sec:theory_conv}.
For strided-convolutions, the im2 family often performs better, whereas, for unstrided convolutions, primitives from the kn2 family are often quicker.
Whereas for unstrided convolutions with specific kernel sizes (one, three and five), families designed for the kernel size (conv-1x1, winograd3, and winograd5) often perform better.

\subsection{Performance Modeling}\label{sec:exp_pm}
We consider a linear regression model (Lin) as baseline and the two neural network models as previously discussed.
The first neural network approach uses a separate fully-connected neural network for each primitive and data-layout transformation (NN1).
The second approach uses only a single fully-connected neural network for all primitives and one for all data-layout transformations (NN2).

The profiler collected dataset is split into three subsets using the 80/10/10 split for training, validation and test respectively.
The initial dataset is shuffled before split.
Performance models are trained on the training set and hyper-parameters adjusted with observations from the validation set.
To choose the neural network architecture size, we iterated by increasing the size until no further performance improvements is observed. 
For the per-primitive model (NN1), a single set of hyper-parameters is used across all models, fine-tuned for the best average performance across all primitives. 
Finally, the best performing model -- as measured on the validation set -- is tested on the test set to assess the final performance.
These specific hyper-parameters we used for the final performance models are presented in Table~\ref{tab:exp_nn_params}.

\begin{table}
	\centering
	\caption{Neural Network Hyper-Parameters. Early stopping was done by only halting the training when the validation performance did not improve for 250 iterations. For fine tuning the learning rate was lowered by a factor of 10.}\label{tab:exp_nn_params}
	\begin{tabular}{@{}lll@{}}
		\toprule
		Setting & NN1 Value & NN2 Value \\
		\midrule
		Optimizer & Adam & Adam \\
		Learning Rate & $0.003$ & $0.001$ \\
		Weight Decay & $0$ & $1\times10^{-5}$ \\
		Batch Size & $1024$ & $1024$ \\
		Iterations & Early Stopping & Early Stopping \\
		Non-Linearity & ReLU & ReLU \\
		Architecture & {\footnotesize$5\times 16 \times 64 \times 64 \times 16 \times 1$} & {\footnotesize$5\times 128 \times 512 \times 512 \times 128 \times n$} \\ 
		\bottomrule
	\end{tabular}
\end{table}

\subsection{Primitive Selection}\label{sec:exp_ps}
We use the best performance model, determined as described in the previous section, to provide the costs for the primitive selection.
To measure the primitive selection performance, we optimise the implementation of six neural networks: Alexnet \cite{alexnet}, VGG-11 and VGG-19 \cite{vgg}, GoogLeNet \cite{googlenet}, and ResNet-18 and ResNet-34 \cite{resnet}.
These are representative networks of current neural networks, with both classic ones (AlexNet) and more wide-spread ones (ResNet).

The six neural networks are optimized with primitive selection on both profiled execution time and on performance model estimations.
The performance model is trained as described in the previous section.
The BPQP-solver is used as an optimiser, and implemented as described in \cite{pbqp}.

\subsection{Transferability}\label{sec:exp_tl}
We train each performance models using the performance measurements from just one platform, which makes them platform dependent. 
It is inconvenient to train the performance models for a new platform from scratch, due to the amount of time it takes to collect all the required training data.
Instead, we apply a transfer learning approach to specialize the performance models to another platform with minimal amount of profiled sample points.

First, we evaluate the effect of applying the performance model trained on the Intel machine directly on the models optimised for the ARM and AMD machines compared to the optimal primitive selection for those target machines.
By this, we run the primitive selection of Intel optimised networks on the other two platforms to see the gap in performance if models are not hardware dependent.
This is tested on the layers of GoogLeNet due to its large variety in convolutional layers.
A biggest degradation in performance may also be due to predictions of the Intel machine being, on average, significantly smaller due to the higher processor clock speed.

Second, we determine a scale by which the estimation of the Intel model can be calibrated to produce more comparable values to those measured on the ARM and AMD platforms.
This scale factor is determined by using only 1\% of the collected performance sample points on the two machines.
Both `factor-corrected' Intel models -- one for the AMD and one for the ARM dataset -- are then evaluated in the primitive selection process.

We know from previous work that by using transfer learning, the burden of collecting a large training set for the target devices is reduced~ \cite{benchmark_ml}.
Here we want to know how much data is actually enough for the fine-tuning stage to produce a quality performance model for a different hardware.
We take the best Intel performance model as the starting point for transfer learning.
For both AMD and ARM platforms, 6 separate fine-tuning stages are performed on fractions of the initial training sets.
These are randomly selected at 0.1\%, 1\%, 2.5\%, 5\%, 10\% and 25\% of the training data available for each of those two devices.
Their estimations are used for the same task as before, as part of the primitive selection process for GoogleNet.
Picking different subsets of the data can give different results.
To evaluate this effect, each model is trained 25 times using a different subset of the data (sampled uniformly at random).
We then compare these results with a performance model trained from scratch -- without transfer learning -- using the same fractions of the data, which we expect to perform worse.

Finally, the transferability between primitive families is evaluated.
This is done by fine-tuning the Intel performance model to the AMD platform using just data from an individual primitive family (such as im2) and evaluating the result on all primitive families separately.
This process is then repeated for all seven primitive families.

\begin{figure*}[t]
	\centering
	\begin{tikzpicture}
\pgfplotstableread{data/mean_nn_err_test10.txt}{\data}
\pgfplotstablegetcolumnnamebyindex{0}\of{\data}\to{\datacol}

\pgfplotstableread{data/std_nn_err_test10.txt}{\error}
\pgfplotstablegetcolumnnamebyindex{0}\of{\error}\to{\errorcol}

\pgfplotstableread{data/mean_nn_full_err_test10.txt}{\datafull}
\pgfplotstablegetcolumnnamebyindex{0}\of{\datafull}\to{\datacol}

\pgfplotstableread{data/std_nn_full_err_test10.txt}{\errorfull}
\pgfplotstablegetcolumnnamebyindex{0}\of{\errorfull}\to{\errorcol}

\pgfplotstableread{data/data_lin_err_adv_test10.txt}{\datalin}
\pgfplotstablegetcolumnnamebyindex{0}\of{\datalin}\to{\datacol}

\pgfplotstableread{data/keys_lin_prim_red.txt}{\keys}

\pgfplotstablecreatecol[copy column from table={\keys}{label}] {label} {\data}
\pgfplotstablecreatecol[copy column from table={\keys}{loc}] {loc} {\data}
\pgfplotstablecreatecol[copy column from table={\error}{\errorcol}] {error} {\data}

\pgfplotstablecreatecol[copy column from table={\keys}{label}] {label} {\datafull}
\pgfplotstablecreatecol[copy column from table={\keys}{loc}] {loc} {\datafull}
\pgfplotstablecreatecol[copy column from table={\errorfull}{\errorcol}] {error} {\datafull}

\pgfplotstablecreatecol[copy column from table={\keys}{label}] {label} {\datalin}
\pgfplotstablecreatecol[copy column from table={\keys}{loc}] {loc} {\datalin}

\begin{axis}[
	ybar,
	scaled y ticks=false,
	ymin=0,
	ybar=0pt,
	ymax=0.20,
	width=1\textwidth,
	height=5cm,
	bar width=2pt,
	enlarge x limits=0.03,
	ylabel={\small MdRAE},
	yticklabel style={font=\footnotesize},
	xticklabel style={font=\tiny\ttfamily},
	xticklabels from table={\data}{label},
	xtick=data,
	ymajorgrids,
	xtick pos=bottom,
	ytick pos=left,
	ytick = {0.0, 0.02, 0.04, 0.06, 0.08, 0.10, 0.12, 0.14, 0.16, 0.18, 0.20},
    yticklabel={\pgfmathparse{\tick*100}\pgfmathprintnumber{\pgfmathresult}\%},
]

\addplot+[blue4,bar shift=0,fill=none] table[x=loc,y=\datacol] {\datalin};
\addlegendentry{\footnotesize Lin}

\addplot+[red4, bar shift=-0.21,
error bars/.cd,
error bar style={red1},
y dir=both,
y explicit,
error mark options={
	mark size=1pt,
	rotate=90,
}] table[x=loc,y=\datacol,y error=error] {\data};
\addlegendentry{\footnotesize NN1}

\addplot+[red3, bar shift=0.21,
error bars/.cd,
error bar style={red1},
y dir=both,
y explicit,
error mark options={
	mark size=1pt,
	rotate=90,
}] table[x=loc,y=\datacol,y error=error] {\datafull};
\addlegendentry{\footnotesize NN2}

\end{axis}
\node at (0.45,-0.6) {\tiny direct};
\node at (2.78,-0.6) {\tiny im2};
\node at (5.78,-0.6) {\tiny kn2};
\node at (8.425,-0.6) {\tiny wino3};
\node at (11.85,-0.6) {\tiny wino5};
\node at (14.475,-0.6) {\tiny conv-1x1};
\node at (15.65,-0.6) {\tiny mec};
\end{tikzpicture}
	\caption{MdRAE (median relative absolute error)
	of primitive execution time predictions with NN1, NN2 and a Linear Regression on the Intel test set.
	A mapping to the specific primitive names can be found in \cref{tab:ap_prims}.
	}
	\label{fig:res_nns}
\end{figure*}
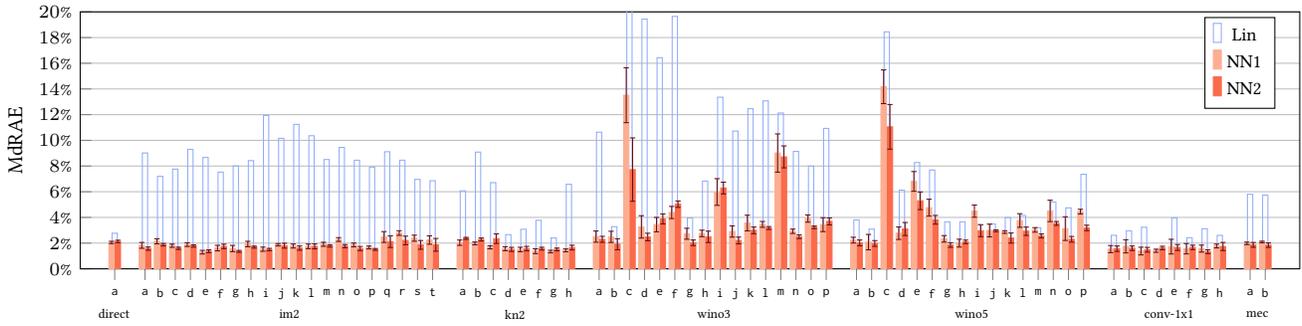

\begin{figure*}[t]
	\centering
	\begin{tikzpicture}
\pgfplotstableread{data/mean_nn_amd_err_test10.txt}{\data}
\pgfplotstablegetcolumnnamebyindex{0}\of{\data}\to{\datacol}

\pgfplotstableread{data/std_nn_amd_err_test10.txt}{\error}
\pgfplotstablegetcolumnnamebyindex{0}\of{\error}\to{\errorcol}

\pgfplotstableread{data/mean_nn_arm_err_test10.txt}{\datafull}
\pgfplotstablegetcolumnnamebyindex{0}\of{\datafull}\to{\datacol}

\pgfplotstableread{data/std_nn_arm_err_test10.txt}{\errorfull}
\pgfplotstablegetcolumnnamebyindex{0}\of{\errorfull}\to{\errorcol}

\pgfplotstableread{data/keys_lin_prim_red.txt}{\keys}

\pgfplotstablecreatecol[copy column from table={\keys}{label}] {label} {\data}
\pgfplotstablecreatecol[copy column from table={\keys}{loc}] {loc} {\data}
\pgfplotstablecreatecol[copy column from table={\error}{\errorcol}] {error} {\data}

\pgfplotstablecreatecol[copy column from table={\keys}{label}] {label} {\datafull}
\pgfplotstablecreatecol[copy column from table={\keys}{loc}] {loc} {\datafull}
\pgfplotstablecreatecol[copy column from table={\errorfull}{\errorcol}] {error} {\datafull}

\begin{axis}[
	ybar,
	scaled y ticks=false,
	ymin=0,
	ybar=0pt,
	ymax=0.15,
	width=1\textwidth,
	height=4.5cm,
	bar width=2pt,
	enlarge x limits=0.03,
	ylabel={\small MdRAE},
	yticklabel style={font=\footnotesize},
	xticklabel style={font=\tiny\ttfamily},
	xticklabels from table={\data}{label},
	xtick=data,
	ymajorgrids,
	xtick pos=bottom,
	ytick pos=left,
	ytick = {0.0, 0.02, 0.04, 0.06, 0.08, 0.10, 0.12, 0.14, 0.16, 0.18, 0.20},
    yticklabel={\pgfmathparse{\tick*100}\pgfmathprintnumber{\pgfmathresult}\%},
]

\addplot+[red4, bar shift=-0.21,
error bars/.cd,
error bar style={red1},
y dir=both,
y explicit,
error mark options={
	mark size=1pt,
	rotate=90,
}] table[x=loc,y=\datacol,y error=error] {\data};
\addlegendentry{\footnotesize AMD}

\addplot+[red3, bar shift=0.21,
error bars/.cd,
error bar style={red1},
y dir=both,
y explicit,
error mark options={
	mark size=1pt,
	rotate=90,
}] table[x=loc,y=\datacol,y error=error] {\datafull};
\addlegendentry{\footnotesize ARM}

\end{axis}
\node at (0.45,-0.6) {\tiny direct};
\node at (2.78,-0.6) {\tiny im2};
\node at (5.78,-0.6) {\tiny kn2};
\node at (8.425,-0.6) {\tiny wino3};
\node at (11.85,-0.6) {\tiny wino5};
\node at (14.475,-0.6) {\tiny conv-1x1};
\node at (15.65,-0.6) {\tiny mec};
\end{tikzpicture}
	\caption{MdRAE (median relative absolute error) 
	of primitive execution time predictions with the NN2 performance model on the AMD and ARM test sets.
	Due to memory constraints on the ARM device, not all primitives could be profiled.
	}
	\label{fig:res_nns_alt}
\end{figure*}
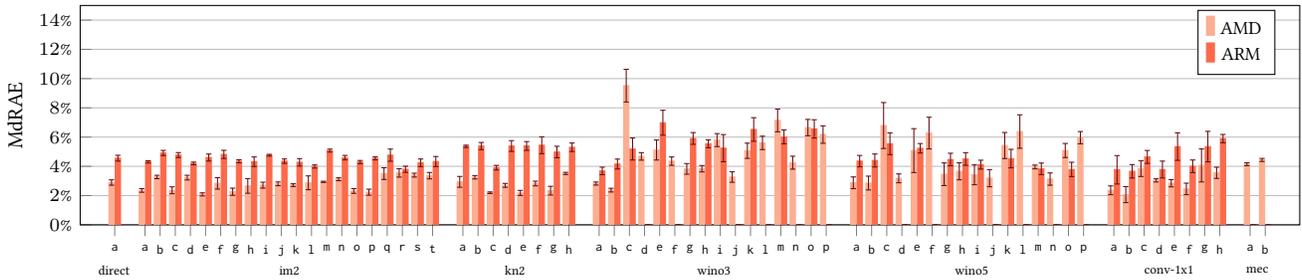

\section{Results and Evaluation}\label{sec:results}
This section presents and discusses the results of our experiments. Here we assess the quality of our performance models, their impact on the primitive selection for CNN optimisation, and the quality and utility of transfer learning for specialising performance models to another target platforms.

\subsection{Performance Model Estimation Accuracy}\label{sec:res_pm}
The evaluation of Lin, NN1 and NN2 performance models trained on the Intel dataset is presented in Figure \ref{fig:res_nns}, expressed in MdRAE over the test set.
Both neural network models significantly outperform the linear regression model on nearly all primitives, justifying the need for non-linear models for performance prediction.
The linear regression solution performs well for some of the primitives, most notably for the direct-sum2d and conv-1x1 primitive families.
This can be explained by the lower complexity of those families of primitives compared to the rest, limited memory requirement and sequential access. 

Except for the winograd family, both neural network models predict all primitives with a MdRAE of around 2\% (i.e. predictions are off by about 2\% in the median values from the profiled execution time on the device for a given layer configuration).
Most primitives in the winograd family perform with a MdRAE between 2\% and 4\%, while some primitives even reach errors as high as 10\%.
The reason why the winograd family may be harder to estimate is likely due to the fewer sample point in the training data for its complexity of performance values.
Between the two neural network models, we see NN2 performing better than NN1 across all primitives. 
This is due to having access to alternative perspectives from all primitives and their performance to draw relations during training. 

We choose the NN2 model as our default performance model to use in the following experiments since it has the lowest MdRAE out of the three models discussed here.
We also validate this observation about our NN2 performance model on the other two platforms -- on the AMD and ARM. 
Similar to the experiments for the Intel platform, we train the NN2 performance model on the profiled data from the AMD and ARM devices. 
The estimation error on their respective test sets is presented in Figure~\ref{fig:res_nns_alt}. 
We see the AMD trained model estimates the performance of primitives about as good as when trained for the Intel platform, just above 2\% estimation MdRAE.
Whereas for the ARM platform the error is in general between 4\% and 6\% for most primitives.

\begin{figure}[t]
	\centering
	\begin{tikzpicture}
\pgfplotstableread{data/mean_nntr_err_test10.txt}{\data}
\pgfplotstablegetcolumnnamebyindex{0}\of{\data}\to{\datacol}

\pgfplotstableread{data/std_nntr_err_test10.txt}{\error}
\pgfplotstablegetcolumnnamebyindex{0}\of{\error}\to{\errorcol}

\pgfplotstableread{data/mean_nntr_full_err_test10.txt}{\datafull}
\pgfplotstablegetcolumnnamebyindex{0}\of{\datafull}\to{\datacol}

\pgfplotstableread{data/std_nntr_full_err_test10.txt}{\errorfull}
\pgfplotstablegetcolumnnamebyindex{0}\of{\errorfull}\to{\errorcol}

\pgfplotstableread{data/data_lintr_err_adv.txt}{\lindata}
\pgfplotstablegetcolumnnamebyindex{0}\of{\lindata}\to{\datacol}

\pgfplotstableread{data/keys_lin_tr_red.txt}{\keys}

\pgfplotstablecreatecol[copy column from table={\keys}{label}] {label} {\data}
\pgfplotstablecreatecol[copy column from table={\keys}{loc}] {loc} {\data}
\pgfplotstablecreatecol[copy column from table={\error}{\errorcol}] {error} {\data}

\pgfplotstablecreatecol[copy column from table={\keys}{label}] {label} {\datafull}
\pgfplotstablecreatecol[copy column from table={\keys}{loc}] {loc} {\datafull}
\pgfplotstablecreatecol[copy column from table={\errorfull}{\errorcol}] {error} {\datafull}

\pgfplotstablecreatecol[copy column from table={\keys}{label}] {label} {\lindata}
\pgfplotstablecreatecol[copy column from table={\keys}{loc}] {loc} {\lindata}

\begin{axis}[
	ybar,
	ybar=0,
	scaled y ticks=false,
	ymin=0,
	ymax=0.05,
	width=0.48\textwidth,
	height=4cm,
	bar width=6pt,
	enlarge x limits=0.10,
	ylabel={\small MdRAE},
	yticklabel style={font=\footnotesize},
	xticklabel style={font=\tiny\ttfamily},
	xticklabels from table={\data}{label},
	xtick=data,
	ymajorgrids,
	xtick pos=bottom,
	ytick pos=left,
	ytick = {0.00, 0.01, 0.02, 0.03, 0.04, 0.05},
    yticklabel={\pgfmathparse{\tick*100}\pgfmathprintnumber{\pgfmathresult}\%},
]

\addplot+[blue4] table[x=loc,y=\datacol] {\lindata};
\addlegendentry{\footnotesize Lin}

\addplot+[red4,
error bars/.cd,
error bar style={red1},
y dir=both,
y explicit] table[x=loc,y=\datacol,y error=error] {\data};
\addlegendentry{\footnotesize NN1}

\addplot+[red3,
error bars/.cd,
error bar style={red1},
y dir=both,
y explicit] table[x=loc,y=\datacol,y error=error] {\datafull};
\addlegendentry{\footnotesize NN2}

\end{axis}
\node at (1.20,-0.2) {\tiny hwc to};
\node at (3.50,-0.2) {\tiny hcw to};
\node at (5.80,-0.2) {\tiny chw to};
\end{tikzpicture}
	\caption{MdRAE -- the median relative absolute error (\cref{sec:ml_ml}) -- for the data layout transformation time predictions of NN1, NN2 and Lin on the test set for the Intel machine.
	The Lin values have an MdRAE of about 10\%.
	}
	\label{fig:res_tr} 
\end{figure}
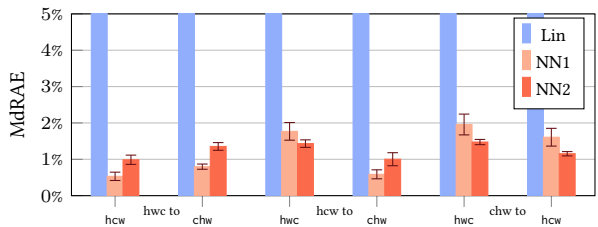

The MdRAE for estimating the execution time of data-layout transformations is presented in Figure \ref{fig:res_tr}.
Transformations from and to the same layout are not presented as these can be skipped from execution (cost zero).
We see the estimations with neural networks are very accurate, most errors being around 1\% for the neural network models. Whereas using the linear regression model we observe very high errors.  
The NN1 models perform better for transformations from `hwc', whereas the NN2 model performs better for transformations from `chw' formats. 
Since NN2 has a relatively stable accuracy across all transformations, around 1\% error, we choose this performance model solution as our default option.

In both cases (for primitive execution time estimation and for data layout transformation time) the NN2 models have a slight edge over the NN1 models. All the following experiments are conducted using the NN2 as our performance models (one for estimating primitive execution time and another one for estimating the execution time of data-layout transformations).

\subsection{Primitive Selection}\label{sec:res_ps}
\paragraph{Selection Speed} Each of the two stages of primitive selection involves a time cost: the time required to measure or estimate the performance (execution time) of primitives/transformations and the time needed to apply the solver across the network layers (PBQP time).  Table \ref{tab:res_times} shows the time required to perform the entire primitives selection on our selected CNNs.
The biggest cost is the time required to profile all the convolutional layers by iterating over the primitives in order to generate the cost graph for the solver.
We can see this is in the range of hours for the smaller platform (ARM), but also significantly larger for the Intel and AMD platforms (0.57 hours and 1.76 hours respectively) when profiling VGG-19.
Performing the optimisation across the network with the PBQP solver incurs its own cost, which is included in the values presented in Table~\ref{tab:res_times}. 
But since this cost is unavoidable for both cases we mainly concentrate on reducing the first cost.

By using the performance model, the speed of estimating the cost of any primitive with any layer configuration is reduced to millisecond.
This is presented in the second column of Table~\ref{tab:res_times} for our chosen networks.
These estimations are produced on the Intel platform.
As observed, estimating the performance of primitives and transformations for the primitive selection of VGG-19 is 673ms, whereas profiling those on the ARM system takes 4.58 hours. 
This indicates a speedup of 25,000$\times$ to optimise the network for the ARM device with our performance model instead of profiling on the device.

\begin{table}[t]
	\centering
	\caption{Total time required to optimize various networks using primitive selection.
	The times are shown for the profiling approach of~\cite{primitive} and our performance model based approach.
	These times include the PBQP-solver time to find the CNN optimal configuration.
	}
	\begin{tabular}{@{}lclll@{}}
		\toprule
		CNN Model & Perf. Model Inf. & \multicolumn{3}{c}{Profiling}\\
		\cmidrule{3-5}
		& & Intel & AMD & Arm \\
		\midrule
		AlexNet & \SI{43.6}{\milli\second} & \SI{66}{\second} & \SI{189}{\second} & \SI{424}{\second} \\
		Vgg11 & \SI{327}{\milli\second} & \SI{0.20}{\hour} & \SI{0.67}{\hour} & \SI{1.72}{\hour} \\
		Vgg19 & \SI{673}{\milli\second} & \SI{0.57}{\hour} & \SI{1.79}{\hour} & \SI{4.58}{\hour} \\
		GoogLeNet & \SI{177}{\milli\second} & \SI{182}{\second} & \SI{445}{\second} & \SI{0.37}{\hour} \\
		ResNet-18 & \SI{119}{\milli\second} & \SI{242}{\second} & \SI{736}{\second} & \SI{0.43}{\hour} \\
		ResNet-34 & \SI{194}{\milli\second} & \SI{494}{\second} & \SI{0.41}{\hour} & \SI{0.89}{\hour} \\
		\bottomrule
	\end{tabular}
	\label{tab:res_times}
\end{table}

\begin{figure}[t]
	\centering
	\begin{tikzpicture}
\begin{axis}[
	ybar,
	ybar=0,
	ymin=0,
	width=0.45\textwidth,
	height=4cm,
	ymax=0.012,
	bar width=6pt,
	enlarge x limits=0.10,
	ylabel={\small Inference Time Increase},
	yticklabel style={font=\footnotesize},
	xticklabel style={font=\tiny\ttfamily},
    symbolic x coords={AlexNet,Vgg11,Vgg19,GNet,RNet-18,RNet-34},
	xtick=data,
	legend pos=north west,
	ymajorgrids,
	ytick = {0, 0.002, 0.004, 0.006, 0.008, 0.01, 0.012},
	scaled y ticks=false,
	xtick pos=bottom,
	ytick pos=left,
	yticklabel={\pgfmathparse{\tick*100}\pgfmathprintnumber{\pgfmathresult}\%},
	tick label style={
		/pgf/number format/fixed,
		/pgf/number format/fixed zerofill,
		/pgf/number format/precision=1,
	}
]

\addplot+[red2] coordinates {(AlexNet,0.0044) (Vgg11,0.0020) (Vgg19,0.0013) (GNet,0.0069) (RNet-18,0.0019) (RNet-34,0.0017)};
\addlegendentry{\footnotesize Intel}

\addplot+[red4] coordinates {(AlexNet,0.0011) (Vgg11,0.0067) (Vgg19,0.0012) (GNet,0.0068) (RNet-18,0.0048) (RNet-34,0.0041)};
\addlegendentry{\footnotesize AMD}

\addplot+[red3] coordinates {(AlexNet,0) (Vgg11,0) (Vgg19,0) (GNet,0.011) (RNet-18,0.0103) (RNet-34,0.0054)};
\addlegendentry{\footnotesize ARM}

\end{axis}

\end{tikzpicture}
	\caption{Relative increase in the inference time when a model is optimized using the costs from the performance model, to the inference time of optimising with the actual measured times. The increase is negligible.
	GNet and RNet denote GoogLeNet and ResNet respectively.}
	\label{fig:res_perf}
\end{figure}
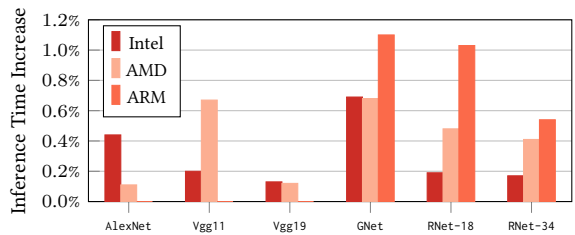

\paragraph{Inference overheads}
After seeing that our performance model can drastically speed-up the primitive selection process, we now analyze the quality of the solution in terms of its inference time for the optimised CNN (with the selected primitives on their estimated performance).

As we have seen, the estimated performance of primitives is very accurate, but the small errors might influence the quality of the identified solution with PBQP. \Cref{fig:res_perf} shows the increase in inference time of networks optimized by PBQP using the NN2 performance model, compared to models optimized using the profiled execution times (i.e. the approach first described in ~\cite{primitive}).
For all networks we considered, the execution time for the optimised network, with estimated performance of primitives, never increased by more than 1.1\%, across all platforms, which is negligible.

Overall, the inference time increase is generally smaller for the Intel platform, less than 0.7\%, and larger for the ARM platform, but still below 1.1\%. In fact, for the ARM platform, the estimated performance helps to find the optimal solution (0\% execution time increased) for three of the networks: AlexNet, VGG-11 and VGG-19. 

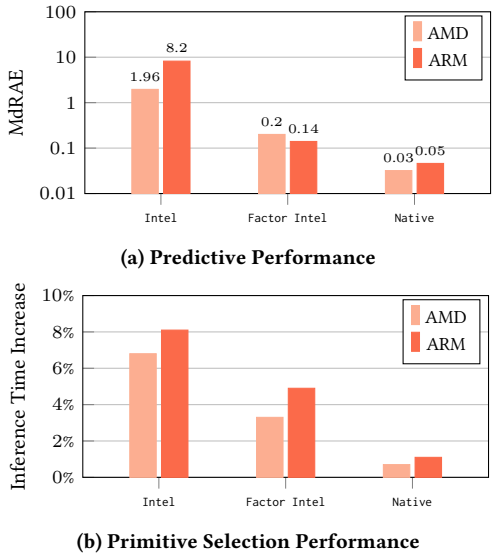
\begin{figure}[t]
	\centering
	\begin{subfigure}{0.49\textwidth}
		\centering
		\begin{tikzpicture}
\begin{axis}[
	ybar,
	ymode=log, 
	log origin=infty,
	width=0.8\textwidth,
	height=4cm,
	ymax=100,
	ymin=0.01,
	every node near coord/.append style={color=black,font=\tiny},
	enlarge x limits=0.3,
	point meta=rawy,
	log ticks with fixed point,
	ylabel={\small MdRAE},
	yticklabel style={font=\footnotesize},
	xticklabel style={font=\tiny\ttfamily},
    symbolic x coords={Intel, Factor Intel, Native},
	xtick=data,
	ymajorgrids,
	scaled y ticks=false,
	xtick pos=bottom,
	ytick pos=left,
	ytick = {100, 10, 1, 0.1, 0.01},
	nodes near coords={\pgfmathprintnumber[fixed, precision=2]{\pgfplotspointmeta}},
]

\addplot+[red4] coordinates {(Intel,1.96) (Factor Intel,0.20) (Native,0.032)};
\addlegendentry{\footnotesize AMD}

\addplot+[red3] coordinates {(Intel,8.20) (Factor Intel,0.14) (Native,0.046)};
\addlegendentry{\footnotesize ARM}

\end{axis}

\end{tikzpicture}
		\caption{Predictive Performance}
		\label{fig:tl_notrain_a}
	\end{subfigure}
	
	\begin{subfigure}{0.49\textwidth}
		\centering
		\begin{tikzpicture}
\begin{axis}[
ybar,
width=0.8\textwidth,
height=4cm,
ymax=0.1,
ymin=0,
every node near coord/.append style={color=black,font=\tiny},
enlarge x limits=0.3,
point meta=rawy,
log ticks with fixed point,
ylabel={\small Inference Time Increase},
yticklabel style={font=\footnotesize},
xticklabel style={font=\tiny\ttfamily},
symbolic x coords={Intel, Factor Intel, Native},
xtick=data,
ymajorgrids,
scaled y ticks=false,
xtick pos=bottom,
ytick pos=left,
yticklabel={\pgfmathparse{\tick*100}\pgfmathprintnumber{\pgfmathresult}\%},
ytick = {0.1, 0.08, 0.06, 0.04, 0.02, 0},
]

\addplot+[red4] coordinates {(Intel,0.068) (Factor Intel,0.033) (Native,0.007)};
\addlegendentry{\footnotesize AMD}

\addplot+[red3] coordinates {(Intel,0.081) (Factor Intel,0.049) (Native,0.011)};
\addlegendentry{\footnotesize ARM}

\end{axis}

\end{tikzpicture}
		\caption{Primitive Selection Performance}
		\label{fig:tl_notrain_b}
	\end{subfigure}
	\caption{
		The accuracy of three approaches to estimate the performance of primitives on the AMD and ARM platforms: (i) Intel - which uses the performance model trained on the Intel platform data directly; (ii) Factor Intel - which uses a small amount of data from the target devices to determine a scale factor for the output of the Intel performance model; (iii) training the performance model directly on the target devices own profiled training dataset.
		This is presented (a) at primitive estimation level in MdRAE and (b) in the CNN inference time as performance estimations influence the solver across the whole network.
	}
	\label{fig:tl_notrain}
\end{figure}

\subsection{Transfer Learning}\label{sec:res_tl}
We first understand the need for transfer learning by observing the gap in performance when adopting the Intel performance model for an ARM and AMD machine -- directly and with output scaling factors. 

\paragraph{Factor correction.} \Cref{fig:tl_notrain_a} shows the MdRAE for the Intel and factor corrected Intel performance model (as described in \cref{sec:exp_tl}) when evaluated on the ARM and AMD test sets.
As expected, applying the Intel performance model directly to a new platform results in a significant reduction in performance.
The MdRAE goes as high as 820\% when evaluated on the ARM dataset, much higher than the native model performance (4.6\%), which is achieved when training the ARM performance mode on the training set of the device.
Just by correcting the outputs of the Intel performance model with a factor (Factor Intel), however, seems to be good enough to adapt the model estimation by a factor to match that of a different target platform. This factor is determined independently for each primitive by using 25 sample points from the device (1\% of the profiled dataset). 
By this approach, the MdRAE of the Intel performance model improves to 14\% when evaluated on the ARM dataset. 

\Cref{fig:tl_notrain_b} shows how using the Intel performance model influences the primitive selection process for a network running on the ARM and AMD platforms (both applied directly and with an output scaling factor). 
These are measured for optimising GoogLeNet with estimated primitive execution time, versus the actual profiled time on the ARM and AMD machines. 
Using uncorrected Intel performance model for the ARM platform (having a MdRAE of 820\%) results in an inference time increase of just over 8\%.
Although this is significantly larger than the native performance (i.e. 1.1\%) when using an ARM performance model, it would still be much quicker than an unoptimized network.
With the factor correction, the inference time increase is half that of the increase when using the Intel performance model directly in optimising the GoogleNet for both ARM and AMD.
We see there is still a gap between these performance estimators and what a NN2 native trained for the AMD and ARM can achieve. 
We set to close this gap by using transfer learning on the Intel trained performance model.

\paragraph{Transfer learning.} \Cref{fig:tl_train} (c) and (d) show how the Intel trained performance model (without factor correction) can be further improved by fine-tuning (i.e. continuing to train the model) using a small fraction of the training data available from profiling the target platform.
By comparison, \Cref{fig:tl_train} (a) and (b) present the accuracy of training the performance model from scratch on the same fraction of training data used for fine-tuning the Intel model.
For both target platforms (i.e. AMD and ARM) 10\% of our original training set is sufficient to train a performance model:
\begin{itemize}
    \item with errors of 7\% and 8\% for training from scratch and 5\% and 5.7\% for transfer learning (predictive performance).
    \item and an inference time increase of 4\% and 5.3\% for training from scratch and 1.4\% and 1.9\% for transfer learning (primitive selection performance).
\end{itemize}
For transfer learning, these are close in performance to a performance model trained over the entire training set (dotted lines).

The differences increase when lowering the amount of data available for training.
The accuracy of the performance model trained from scratch degrades steeply when using just 1\% of the original training data.  
The inference time increases by more than 20\% when using the performance model trained from scratch.
Whereas the transfer learning model performs much better, increasing the configured network inference time by only 4\%.
\Cref{fig:tl_vs_scratch_0001} shows an even larger difference when using only 0.1\% of the training data.
At the other extreme, making more data available for training (more than 10\%), benefits both the trained from scratch model and the transfer learning model.
At 25\% of data, our transfer learning performance model is only marginally worse (less than 1\% error) than a model trained on all available data (\Cref{fig:tl_train}).

\begin{table}[t]
	\centering
	\caption{The (relative) predictive performance of the Intel performance model fine-tuned to the AMD platform. The model is fine-tuned using only data from one primitive family. The columns show on which primitive family the model is evaluated, the rows show which primitive family was used for training. The rows are normalized such to set the diagonal to one (which signifies model trained and evaluated on the same primitive family).}
	\pgfplotstabletypeset[color cells]{
	x, direct,im2,kn2,wino3,wino5,c1x1,mec
	direct, 1, 44, 32, 28, 28, 26, 18 
	im2, 3, 1, 6, 8, 8, 3, 3 
	kn2, 2, 13, 1, 10, 8, 8, 4 
	wino3, 5, 18, 11, 1, 3, 3, 4
	wino5, 6, 16, 10, 4, 1, 4, 5 
	c1x1, 8, 32, 24, 11, 15, 1, 12 
	mec, 5, 36, 24, 23, 21, 19, 1
}
	\label{fig:tl_prim_fam_mat}
\end{table}

\begin{figure*}[t]
	\centering
	\begin{subfigure}[t]{0.45\textwidth}
		\centering
		\begin{tikzpicture}
\begin{axis}[
ymax=0.2,
xmin=0,
ymin=0,
xmax=0.26,
width=1\textwidth,
height=0.65\textwidth,
legend pos=north east,
legend cell align={left},
ymajorgrids,
xmajorgrids,
xtick pos=bottom,
ytick pos=left,
ylabel={MdRAE},
xlabel={Data Usage},
yticklabel={\pgfmathparse{\tick*100}\pgfmathprintnumber{\pgfmathresult}\%},
scaled y ticks=false,
tick label style={
	/pgf/number format/fixed,
}
]

\addplot+[blue4,thick,
mark=none,
error bars/.cd,
error bar style={blue1},
y dir=both,
y explicit] table[x=data,y=mean,y error=std]{data/scratch_exp_new.txt};
\addlegendentry{\footnotesize AMD}

\addplot+[blue2,thick,
mark=none,
error bars/.cd,
error bar style={blue1},
y dir=both,
y explicit] table[x=data,y=mean,y error=std]{data/scratch_exp_new_arm.txt};
\addlegendentry{\footnotesize ARM}

\addplot+[blue4,thick,dashed,mark=none] coordinates {(-1,0.032) (2,0.032)};
\addplot+[blue2,thick,dashed,mark=none] coordinates {(-1,0.046) (2,0.046)};

\end{axis}
\end{tikzpicture}
		\caption{Predictive Performance (trained from Scratch)}
	\end{subfigure}
	\begin{subfigure}[t]{0.45\textwidth}
		\centering
		\begin{tikzpicture}
\begin{axis}[
ymax=0.3,
ymin=0,
xmin=0,
xmax=0.26,
width=1\textwidth,
height=0.65\textwidth,
legend pos=north east,
legend cell align={left},
ymajorgrids,
xmajorgrids,
xtick pos=bottom,
ytick pos=left,
ylabel={Inference Time Increase},
xlabel={Data Usage},
yticklabel={\pgfmathparse{\tick*100}\pgfmathprintnumber{\pgfmathresult}\%},
ytick ={0,0.05,0.1,0.15,0.2,0.25,0.3},
scaled y ticks=false,
tick label style={
	/pgf/number format/fixed,
}
]

\addplot+[blue4,thick,
mark=none,
error bars/.cd,
error bar style={blue1},
y dir=both,
y explicit] table[x=data,y=mean,y error=std]{data/scratch_exp_psel_new.txt};
\addlegendentry{\footnotesize AMD}

\addplot+[blue2,thick,
mark=none,
error bars/.cd,
error bar style={blue1},
y dir=both,
y explicit] table[x=data,y=mean,y error=std]{data/scratch_exp_psel_arm_new.txt};
\addlegendentry{\footnotesize ARM}

\addplot+[blue4,thick,dashed,mark=none] coordinates {(-1,0.0068) (2,0.0068)};
\addplot+[blue2,thick,dashed,mark=none] coordinates {(-1,0.011) (2,0.011)};

\end{axis}
\end{tikzpicture}
		\caption{Primitive Selection Performance (trained from Scratch)}
	\end{subfigure}
	
	\begin{subfigure}[t]{0.45\textwidth}
		\centering
		\begin{tikzpicture}
\begin{axis}[
ymax=0.1,
xmin=0,
ymin=0,
xmax=0.26,
width=1\textwidth,
height=0.65\textwidth,
legend pos=north east,
legend cell align={left},
ymajorgrids,
xmajorgrids,
xtick pos=bottom,
ytick pos=left,
ylabel={MdRAE},
xlabel={Data Usage},
yticklabel={\pgfmathparse{\tick*100}\pgfmathprintnumber{\pgfmathresult}\%},
scaled y ticks=false,
tick label style={
	/pgf/number format/fixed,
}
]

\addplot+[red4,thick,
mark=none,
error bars/.cd,
error bar style={red1},
y dir=both,
y explicit] table[x=data,y=mean,y error=std]{data/tl_exp_new.txt};
\addlegendentry{\footnotesize AMD}

\addplot+[red2,thick,
mark=none,
error bars/.cd,
error bar style={red1},
y dir=both,
y explicit] table[x=data,y=mean,y error=std]{data/tl_exp_new_arm.txt};
\addlegendentry{\footnotesize ARM}

\addplot+[blue4,thick,dashed,mark=none] coordinates {(-1,0.032) (2,0.032)};
\addplot+[blue2,thick,dashed,mark=none] coordinates {(-1,0.046) (2,0.046)};

\end{axis}
\end{tikzpicture}
		\caption{Predictive Performance (using TL)}
	\end{subfigure}
	\begin{subfigure}[t]{0.45\textwidth}
		\centering
		\begin{tikzpicture}
\begin{axis}[
ymax=0.05,
ymin=0,
xmin=0,
xmax=0.26,
width=1\textwidth,
height=0.65\textwidth,
legend pos=north east,
legend cell align={left},
ymajorgrids,
xmajorgrids,
xtick pos=bottom,
ytick pos=left,
ylabel={Inference Time Increase},
xlabel={Data Usage},
yticklabel={\pgfmathparse{\tick*100}\pgfmathprintnumber{\pgfmathresult}\%},
ytick ={0,0.01,0.02,0.03,0.04,0.05},
scaled y ticks=false,
tick label style={
	/pgf/number format/fixed,
}
]

\addplot+[red4,thick,
mark=none,
error bars/.cd,
error bar style={red1},
y dir=both,
y explicit] table[x=data,y=mean,y error=std]{data/tl_exp_psel_new.txt};
\addlegendentry{\footnotesize AMD}

\addplot+[red2,thick,
mark=none,
error bars/.cd,
error bar style={red1},
y dir=both,
y explicit] table[x=data,y=mean,y error=std]{data/tl_exp_psel_arm_new.txt};
\addlegendentry{\footnotesize ARM}

\addplot+[blue4,thick,dashed,mark=none] coordinates {(-1,0.0068) (2,0.0068)};
\addplot+[blue2,thick,dashed,mark=none] coordinates {(-1,0.011) (2,0.011)};

\end{axis}
\end{tikzpicture}
		\caption{Primitive Selection Performance (using TL)}
	\end{subfigure}
	\caption{
	    The predictive and primitive selection performance of the ARM and AMD performance models, trained using various amounts of training data (shown on the x-axis).
	    In the top two figures, the models are trained from scratch with the indicated amount of data from the original training set. In the bottom two, the Intel performance model was used as a starting point to fine-tune with that data. 
	    In all figures, the dotted lines shows the performance of the ARM and AMD performance models when trained from scratch using all the available training data.
	}
	\vspace{-0.2cm}
	\label{fig:tl_train}
\end{figure*}
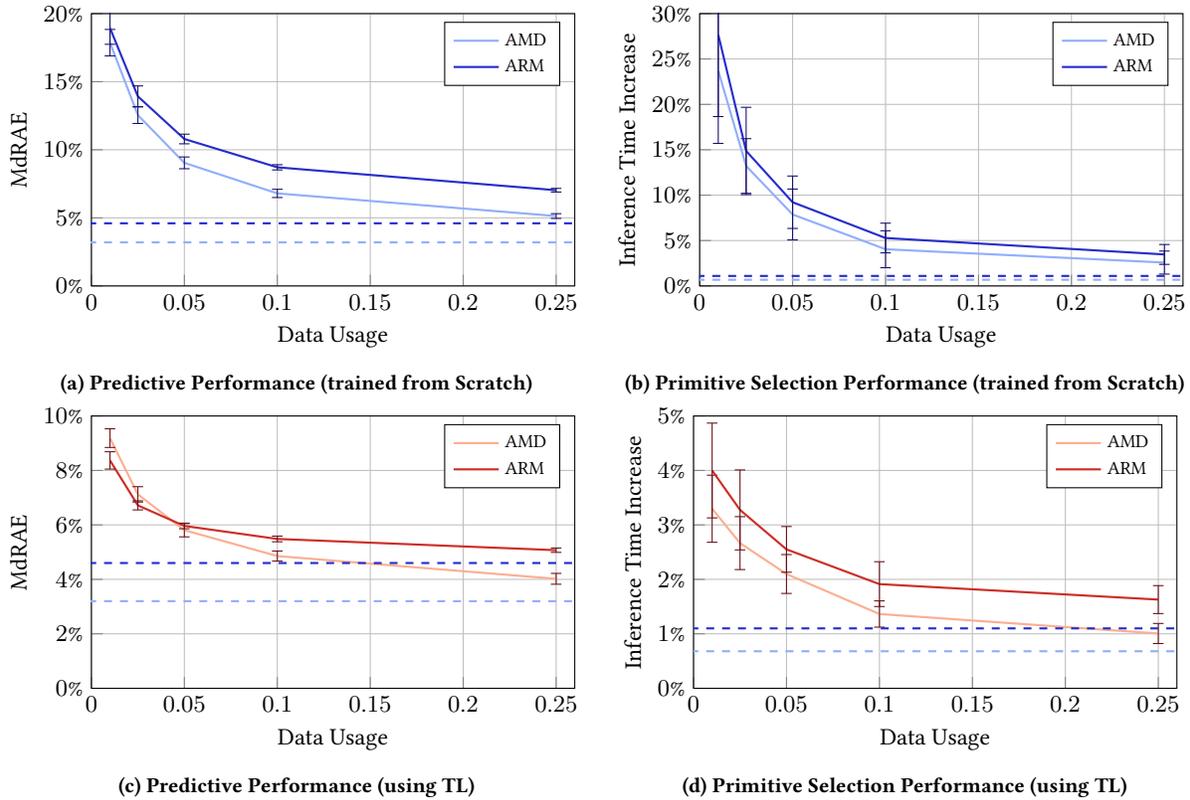

\begin{figure*}[t]
	\centering
	\begin{subfigure}{0.48\textwidth}
		\centering
		\begin{tikzpicture}
\begin{axis}[
	ybar,
	width=0.8\textwidth,
	height=4cm,
	ymax=2,
	ymin=0,
	every node near coord/.append style={color=black,font=\tiny},
	enlarge x limits=1,
	point meta=rawy,
	ylabel={\small MdRAE},
	yticklabel style={font=\footnotesize},
	xticklabel style={font=\tiny\ttfamily},
    symbolic x coords={AMD, ARM},
	xtick=data,
	ymajorgrids,
	scaled y ticks=false,
	xtick pos=bottom,
	ytick pos=left,
	yticklabel={\pgfmathparse{\tick*100}\pgfmathprintnumber{\pgfmathresult}\%},
	ytick = {0, 0.5, 1, 1.5, 2},
]

\addplot+[blue4] coordinates {(AMD,1.328) (ARM,1.625)};
\addlegendentry{\footnotesize Scratch}

\addplot+[red2] coordinates {(AMD,0.177) (ARM,0.269) };
\addlegendentry{\footnotesize TL}

\end{axis}

\end{tikzpicture}
		\caption{Predictive Performance}
		\label{fig:tl_vs_scratch_0001_pred}
	\end{subfigure}
	\begin{subfigure}{0.48\textwidth}
		\centering
		\begin{tikzpicture}
\begin{axis}[
	ybar,
	width=0.8\textwidth,
	height=4cm,
	ymax=1,
	ymin=0,
	every node near coord/.append style={color=black,font=\tiny},
	enlarge x limits=1,
	point meta=rawy,
	log ticks with fixed point,
	ylabel={\small Inference Time Increase},
	yticklabel style={font=\footnotesize},
	xticklabel style={font=\tiny\ttfamily},
	symbolic x coords={AMD, ARM},
	xtick=data,
	ymajorgrids,
	scaled y ticks=false,
	xtick pos=bottom,
	ytick pos=left,
	yticklabel={\pgfmathparse{\tick*100}\pgfmathprintnumber{\pgfmathresult}\%},
	ytick = {1, 0.8, 0.6, 0.4, 0.2, 0},
]

\addplot+[blue4] coordinates {(AMD,0.636) (ARM,0.707)};
\addlegendentry{\footnotesize Scratch}

\addplot+[red2] coordinates {(AMD,0.043) (ARM,0.051)};
\addlegendentry{\footnotesize TL}

\end{axis}

\end{tikzpicture}
		\caption{Primitive Selection Performance}
		\label{fig:tl_vs_scratch_0001_psel}
	\end{subfigure}
	\caption{
	    The predictive and primitive selection performance of the ARM and AMD performance models, trained using 0.1\% of the data.
		The performance of models trained from scratch is compared to models trained using transfer learning (using the Intel performance model as a starting point).
		This figure is an extension to \cref{fig:tl_train} (which shows the same comparison for different amounts of data).
	}
	\label{fig:tl_vs_scratch_0001}
\end{figure*}
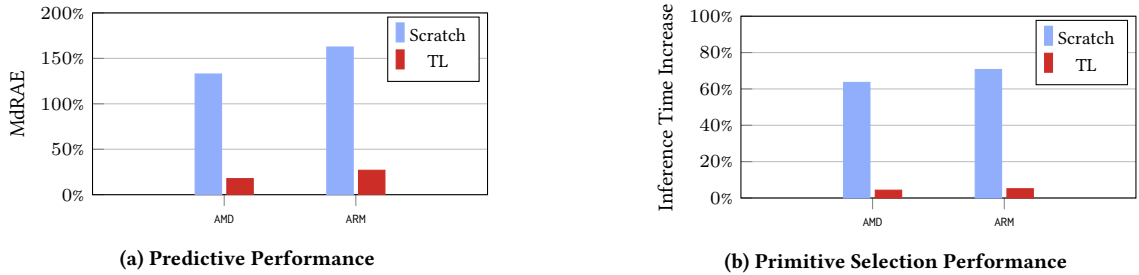

Finally, \cref{fig:tl_prim_fam_mat} provides further insight into how well data from one primitive family transfers to the other families.
From this table, we can see, for example, that when the Intel performance model is fine-tuned with just data from the wino3 primitive family, it still performs reasonably well on wino5 primitives, but performs badly on the im2 and kn2 family.
A model trained with just data on the im2 primitive family, on the other hand, performs reasonably well on all other primitive families.
Beyond learning from just one family of primitives, specialising the model even further requires some data from other families, but in lower amounts.
As seen, the model already performs well when using just data from the im2 primitives.
Supplementing that with a smalls number of data points from the other families can result in a model almost as good as the one trained on all our profiled training data.

\section{Conclusions}\label{sec:conclusion}
We introduce primitive inference time modeling with a neural network as an efficient solution for removing the heavy profiling in primitive selection.
Our approach reduces the time taken for obtaining the best configuration of a neural network from hours to seconds. 
This improvement enables any unseen CNN to quickly reach high performance on any given hardware platform.
We show that the inference time of our configured networks increases by only 0.39\% on average (worse case 1.1\%) relative to the lengthy profiling approach in primitive selection.
Our performance model is also viable for transfer learning, requiring only a small amount of profiling on a new platform to accurately adjust the pre-trained performance model.
We lower the amount of training data required to construct a performance model from scratch to just 1\% when using transfer learning, which results in a below 4\% increase in inference time for our configured network. 

\paragraph{Future Work}
Here, we explored primitive performance modeling only on CPUs. 
The next step is to expand to performance modeling of primitives on GPUs.
This will enable heterogneous CPU and GPU primitive selection.
A `data-transfer' cost (similar to that of data-layout transformations) will be used when transferring data between devices.
	
\Cref{sec:res_tl} showed that transfer learning can vastly reduce the required amount of data.
The particular subset of relevant training points needed in fine-tuning can be determined to minimize the required profiled configuration points, while maximizing the performance.

\balance

\bibliographystyle{ACM-Reference-Format}
\bibliography{bib.bib}


\begin{thebibliography}{31}


\ifx \showCODEN    \undefined \def \showCODEN     #1{\unskip}     \fi
\ifx \showDOI      \undefined \def \showDOI       #1{#1}\fi
\ifx \showISBNx    \undefined \def \showISBNx     #1{\unskip}     \fi
\ifx \showISBNxiii \undefined \def \showISBNxiii  #1{\unskip}     \fi
\ifx \showISSN     \undefined \def \showISSN      #1{\unskip}     \fi
\ifx \showLCCN     \undefined \def \showLCCN      #1{\unskip}     \fi
\ifx \shownote     \undefined \def \shownote      #1{#1}          \fi
\ifx \showarticletitle \undefined \def \showarticletitle #1{#1}   \fi
\ifx \showURL      \undefined \def \showURL       {\relax}        \fi
\providecommand\bibfield[2]{#2}
\providecommand\bibinfo[2]{#2}
\providecommand\natexlab[1]{#1}
\providecommand\showeprint[2][]{arXiv:#2}

\bibitem[\protect\citeauthoryear{Anderson and Gregg}{Anderson and
  Gregg}{2018}]%
        {primitive}
\bibfield{author}{\bibinfo{person}{Andrew Anderson} {and}
  \bibinfo{person}{David Gregg}.} \bibinfo{year}{2018}\natexlab{}.
\newblock \showarticletitle{Optimal DNN Primitive Selection with Partitioned
  Boolean Quadratic Programming}. In \bibinfo{booktitle}{\emph{Proceedings of
  the 2018 International Symposium on Code Generation and Optimization}}.
  \bibinfo{pages}{340--351}.
\newblock


\bibitem[\protect\citeauthoryear{Anderson, Vasudevan, Keane, and
  Gregg}{Anderson et~al\mbox{.}}{2017}]%
        {im2var}
\bibfield{author}{\bibinfo{person}{Andrew Anderson}, \bibinfo{person}{Aravind
  Vasudevan}, \bibinfo{person}{Cormac Keane}, {and} \bibinfo{person}{David
  Gregg}.} \bibinfo{year}{2017}\natexlab{}.
\newblock \showarticletitle{Low-memory GEMM-based convolution algorithmsfor
  deep neural networks}.
\newblock \bibinfo{journal}{\emph{ArXiv}}  \bibinfo{volume}{1709.03395}
  (\bibinfo{year}{2017}).
\newblock


\bibitem[\protect\citeauthoryear{Blahut}{Blahut}{2010}]%
        {quickwino}
\bibfield{author}{\bibinfo{person}{Richard~E. Blahut}.}
  \bibinfo{year}{2010}\natexlab{}.
\newblock \bibinfo{booktitle}{\emph{Fast Algorithms for Signal Processing}}.
\newblock \bibinfo{publisher}{Cambridge University Press}.
\newblock


\bibitem[\protect\citeauthoryear{Cai, Juan, Stamoulis, and Marculescu}{Cai
  et~al\mbox{.}}{2017}]%
        {cnn_power_runtime_pred}
\bibfield{author}{\bibinfo{person}{Ermao Cai}, \bibinfo{person}{Da-Cheng Juan},
  \bibinfo{person}{Dimitrios Stamoulis}, {and} \bibinfo{person}{Diana
  Marculescu}.} \bibinfo{year}{2017}\natexlab{}.
\newblock \showarticletitle{NeuralPower: Predict and Deploy Energy-Efficient
  Convolutional Neural Networks}. In \bibinfo{booktitle}{\emph{Proceedings of
  Machine Learning Research}}. \bibinfo{pages}{622--637}.
\newblock


\bibitem[\protect\citeauthoryear{Colangelo, Nasiri, Nurvitadhi, Mishra,
  Margala, and Nealis}{Colangelo et~al\mbox{.}}{2018}]%
        {low_float}
\bibfield{author}{\bibinfo{person}{Philip Colangelo}, \bibinfo{person}{Nasibeh
  Nasiri}, \bibinfo{person}{Eriko Nurvitadhi}, \bibinfo{person}{Asit Mishra},
  \bibinfo{person}{Martin Margala}, {and} \bibinfo{person}{Kevin Nealis}.}
  \bibinfo{year}{2018}\natexlab{}.
\newblock \showarticletitle{Exploration of Low Numeric Precision Deep Learning
  Inference Using Intel FPGAs}. In \bibinfo{booktitle}{\emph{Proceedings of the
  2018 ACM/SIGDA International Symposium on Field-Programmable Gate Arrays}}.
\newblock


\bibitem[\protect\citeauthoryear{de~Prado, Pazos, and Benini}{de~Prado
  et~al\mbox{.}}{2019}]%
        {primitive_rl}
\bibfield{author}{\bibinfo{person}{Miguel de Prado}, \bibinfo{person}{Nuria
  Pazos}, {and} \bibinfo{person}{Luca Benini}.}
  \bibinfo{year}{2019}\natexlab{}.
\newblock \showarticletitle{Learning to infer: RL-based search for DNN
  primitive selection on Heterogeneous Embedded Systems}. In
  \bibinfo{booktitle}{\emph{2019 Design, Automation Test in Europe Conference
  Exhibition (DATE)}}. \bibinfo{pages}{1409--1414}.
\newblock


\bibitem[\protect\citeauthoryear{Deng, Dong, Socher, Li, Li, and Fei-Fei}{Deng
  et~al\mbox{.}}{2009}]%
        {imagenet}
\bibfield{author}{\bibinfo{person}{J. Deng}, \bibinfo{person}{W. Dong},
  \bibinfo{person}{R. Socher}, \bibinfo{person}{L.-J. Li}, \bibinfo{person}{K.
  Li}, {and} \bibinfo{person}{L. Fei-Fei}.} \bibinfo{year}{2009}\natexlab{}.
\newblock \showarticletitle{{ImageNet: A Large-Scale Hierarchical Image
  Database}}. In \bibinfo{booktitle}{\emph{CVPR09}}. \bibinfo{pages}{248--255}.
\newblock


\bibitem[\protect\citeauthoryear{Dong, Cheng, Juan, Wei, and Sun}{Dong
  et~al\mbox{.}}{2018}]%
        {pppnet}
\bibfield{author}{\bibinfo{person}{Jin-Dong Dong}, \bibinfo{person}{An-Chieh
  Cheng}, \bibinfo{person}{Da-Cheng Juan}, \bibinfo{person}{Wei Wei}, {and}
  \bibinfo{person}{Min Sun}.} \bibinfo{year}{2018}\natexlab{}.
\newblock \showarticletitle{PPP-Net: Platform-aware Progressive Search for
  Pareto-optimal Neural Architectures}. In \bibinfo{booktitle}{\emph{ICLR
  Workshop}}.
\newblock


\bibitem[\protect\citeauthoryear{Hames and Scholz}{Hames and Scholz}{2006}]%
        {pbqp}
\bibfield{author}{\bibinfo{person}{Lang Hames} {and} \bibinfo{person}{Bernhard
  Scholz}.} \bibinfo{year}{2006}\natexlab{}.
\newblock \showarticletitle{Nearly Optimal Register Allocation with PBQP}. In
  \bibinfo{booktitle}{\emph{Proceedings of the 7th Joint Modular Languages
  Conference (JMLC’06). LNCS}}. \bibinfo{pages}{346--361}.
\newblock


\bibitem[\protect\citeauthoryear{He, Zhang, Ren, and Sun}{He
  et~al\mbox{.}}{2016}]%
        {resnet}
\bibfield{author}{\bibinfo{person}{Kaiming He}, \bibinfo{person}{Xiangyu
  Zhang}, \bibinfo{person}{Shaoqing Ren}, {and} \bibinfo{person}{Jian Sun}.}
  \bibinfo{year}{2016}\natexlab{}.
\newblock \showarticletitle{Deep Residual Learning for Image Recognition}. In
  \bibinfo{booktitle}{\emph{2016 IEEE Conference on Computer Vision and Pattern
  Recognition (CVPR)}}. \bibinfo{pages}{770--778}.
\newblock


\bibitem[\protect\citeauthoryear{Howard, Zhu, Chen, Kalenichenko, Wang, Weyand,
  Andreetto, and Adam}{Howard et~al\mbox{.}}{2017}]%
        {mobilenets}
\bibfield{author}{\bibinfo{person}{Andrew~G. Howard}, \bibinfo{person}{Menglong
  Zhu}, \bibinfo{person}{Bo Chen}, \bibinfo{person}{Dmitry Kalenichenko},
  \bibinfo{person}{Weijun Wang}, \bibinfo{person}{Tobias Weyand},
  \bibinfo{person}{Marco Andreetto}, {and} \bibinfo{person}{Hartwig Adam}.}
  \bibinfo{year}{2017}\natexlab{}.
\newblock \showarticletitle{MobileNets: Efficient Convolutional Neural Networks
  for Mobile Vision Applications}.
\newblock \bibinfo{journal}{\emph{ArXiv}}  \bibinfo{volume}{1704.04861}
  (\bibinfo{year}{2017}).
\newblock


\bibitem[\protect\citeauthoryear{Hsu, Chang, Juan, Pan, Chen, Wei, and
  Chang}{Hsu et~al\mbox{.}}{2018}]%
        {monas}
\bibfield{author}{\bibinfo{person}{Chi-Hung Hsu}, \bibinfo{person}{Shu-Huan
  Chang}, \bibinfo{person}{Da-Cheng Juan}, \bibinfo{person}{Jia-Yu Pan},
  \bibinfo{person}{Yu-Ting Chen}, \bibinfo{person}{Wenli Wei}, {and}
  \bibinfo{person}{Shih-Chieh Chang}.} \bibinfo{year}{2018}\natexlab{}.
\newblock \showarticletitle{MONAS: Multi-Objective Neural Architecture Search
  using Reinforcement Learning}.
\newblock \bibinfo{journal}{\emph{ArXiv}}  \bibinfo{volume}{1806.10332}
  (\bibinfo{year}{2018}).
\newblock


\bibitem[\protect\citeauthoryear{{Huang}, {Liu}, v.~d. {Maaten}, and
  {Weinberger}}{{Huang} et~al\mbox{.}}{2017}]%
        {densenet}
\bibfield{author}{\bibinfo{person}{G. {Huang}}, \bibinfo{person}{Z. {Liu}},
  \bibinfo{person}{L. v.~d. {Maaten}}, {and} \bibinfo{person}{K.~Q.
  {Weinberger}}.} \bibinfo{year}{2017}\natexlab{}.
\newblock \showarticletitle{Densely Connected Convolutional Networks}. In
  \bibinfo{booktitle}{\emph{2017 IEEE Conference on Computer Vision and Pattern
  Recognition (CVPR)}}. \bibinfo{pages}{2261--2269}.
\newblock


\bibitem[\protect\citeauthoryear{Hutter, Xu, Hoos, and Leyton-Brown}{Hutter
  et~al\mbox{.}}{2014}]%
        {runtime_pred_alogs}
\bibfield{author}{\bibinfo{person}{Frank Hutter}, \bibinfo{person}{Lin Xu},
  \bibinfo{person}{Holger~H. Hoos}, {and} \bibinfo{person}{Kevin
  Leyton-Brown}.} \bibinfo{year}{2014}\natexlab{}.
\newblock \showarticletitle{Algorithm runtime prediction: Methods \&
  evaluation}.
\newblock \bibinfo{journal}{\emph{Artificial Intelligence}}
  \bibinfo{volume}{206} (\bibinfo{year}{2014}), \bibinfo{pages}{79--111}.
\newblock


\bibitem[\protect\citeauthoryear{Iandola, Moskewicz, Ashraf, Han, Dally, and
  Keutzer}{Iandola et~al\mbox{.}}{2017}]%
        {squeezenet}
\bibfield{author}{\bibinfo{person}{Forrest~N. Iandola},
  \bibinfo{person}{Matthew~W. Moskewicz}, \bibinfo{person}{Khalid Ashraf},
  \bibinfo{person}{Song Han}, \bibinfo{person}{William~J. Dally}, {and}
  \bibinfo{person}{Kurt Keutzer}.} \bibinfo{year}{2017}\natexlab{}.
\newblock \showarticletitle{SqueezeNet: AlexNet-level accuracy with 50x fewer
  parameters and <1MB model size}.
\newblock \bibinfo{journal}{\emph{ArXiv}}  \bibinfo{volume}{1602.07360}
  (\bibinfo{year}{2017}).
\newblock


\bibitem[\protect\citeauthoryear{Jia}{Jia}{2014}]%
        {im2}
\bibfield{author}{\bibinfo{person}{Yangqing Jia}.}
  \bibinfo{year}{2014}\natexlab{}.
\newblock \emph{\bibinfo{title}{Learning Semantic Image Representations at a
  Large Scale}}.
\newblock \bibinfo{thesistype}{Ph.D. Dissertation}. \bibinfo{school}{EECS
  Department, University of California, Berkeley}.
\newblock


\bibitem[\protect\citeauthoryear{Justus, Brennan, Bonner, and McGough}{Justus
  et~al\mbox{.}}{2018}]%
        {cnn_cost_prediction}
\bibfield{author}{\bibinfo{person}{Daniel Justus}, \bibinfo{person}{John
  Brennan}, \bibinfo{person}{Stephen Bonner}, {and} \bibinfo{person}{A.~Stephen
  McGough}.} \bibinfo{year}{2018}\natexlab{}.
\newblock \showarticletitle{Predicting the Computational Cost of Deep Learning
  Models}.
\newblock \bibinfo{journal}{\emph{2018 IEEE International Conference on Big
  Data (Big Data)}} (\bibinfo{year}{2018}), \bibinfo{pages}{3873--3882}.
\newblock


\bibitem[\protect\citeauthoryear{Kim, Reddy, Yun, and Seo}{Kim
  et~al\mbox{.}}{2017}]%
        {nemo}
\bibfield{author}{\bibinfo{person}{Ye-Hoon Kim}, \bibinfo{person}{Bhargava
  Reddy}, \bibinfo{person}{Sojung Yun}, {and} \bibinfo{person}{Chanwon Seo}.}
  \bibinfo{year}{2017}\natexlab{}.
\newblock \showarticletitle{NEMO : Neuro-Evolution with Multiobjective
  Optimization of Deep Neural Network for Speed and Accuracy}. In
  \bibinfo{booktitle}{\emph{Proceedings of the Genetic and Evolutionary
  Computation Conference}}. \bibinfo{pages}{419--427}.
\newblock


\bibitem[\protect\citeauthoryear{Krizhevsky, Sutskever, and Hinton}{Krizhevsky
  et~al\mbox{.}}{2012}]%
        {alexnet}
\bibfield{author}{\bibinfo{person}{Alex Krizhevsky}, \bibinfo{person}{Ilya
  Sutskever}, {and} \bibinfo{person}{Geoffrey~E Hinton}.}
  \bibinfo{year}{2012}\natexlab{}.
\newblock \showarticletitle{ImageNet Classification with Deep Convolutional
  Neural Networks}. In \bibinfo{booktitle}{\emph{Advances in Neural Information
  Processing Systems 25}}. \bibinfo{pages}{1097--1105}.
\newblock


\bibitem[\protect\citeauthoryear{Malakar, Balaprakash, Vishwanath, Morozov, and
  Kumaran}{Malakar et~al\mbox{.}}{2018}]%
        {benchmark_ml}
\bibfield{author}{\bibinfo{person}{Preeti Malakar}, \bibinfo{person}{Prasanna
  Balaprakash}, \bibinfo{person}{Venkatram Vishwanath}, \bibinfo{person}{Vitali
  Morozov}, {and} \bibinfo{person}{Kalyan Kumaran}.}
  \bibinfo{year}{2018}\natexlab{}.
\newblock \showarticletitle{Benchmarking Machine Learning Methods for
  Performance Modeling of Scientific Applications}. In
  \bibinfo{booktitle}{\emph{2018 IEEE/ACM Performance Modeling, Benchmarking
  and Simulation of High Performance Computer Systems (PMBS)}}.
  \bibinfo{pages}{33--44}.
\newblock


\bibitem[\protect\citeauthoryear{Radu, Kaszyk, Wen, Turner, Cano, Crowley,
  Franke, Storkey, and O’Boyle}{Radu et~al\mbox{.}}{2019}]%
        {radu2019performance}
\bibfield{author}{\bibinfo{person}{Valentin Radu}, \bibinfo{person}{Kuba
  Kaszyk}, \bibinfo{person}{Yuan Wen}, \bibinfo{person}{Jack Turner},
  \bibinfo{person}{Jos{\'e} Cano}, \bibinfo{person}{Elliot~J Crowley},
  \bibinfo{person}{Bj{\"o}rn Franke}, \bibinfo{person}{Amos Storkey}, {and}
  \bibinfo{person}{Michael O’Boyle}.} \bibinfo{year}{2019}\natexlab{}.
\newblock \showarticletitle{Performance Aware Convolutional Neural Network
  Channel Pruning for Embedded GPUs}.
\newblock  (\bibinfo{year}{2019}).
\newblock


\bibitem[\protect\citeauthoryear{Simonyan and Zisserman}{Simonyan and
  Zisserman}{2014}]%
        {vgg}
\bibfield{author}{\bibinfo{person}{Karen Simonyan} {and}
  \bibinfo{person}{Andrew Zisserman}.} \bibinfo{year}{2014}\natexlab{}.
\newblock \showarticletitle{Very Deep Convolutional Networks for Large-Scale
  Image Recognition}.
\newblock \bibinfo{journal}{\emph{CoRR}}  \bibinfo{volume}{1409.1556}
  (\bibinfo{year}{2014}).
\newblock


\bibitem[\protect\citeauthoryear{Szegedy, Liu, Jia, Sermanet, Reed, Anguelov,
  Erhan, Vanhoucke, and Rabinovich}{Szegedy et~al\mbox{.}}{2015}]%
        {googlenet}
\bibfield{author}{\bibinfo{person}{Christian Szegedy}, \bibinfo{person}{Wei
  Liu}, \bibinfo{person}{Yangqing Jia}, \bibinfo{person}{Pierre Sermanet},
  \bibinfo{person}{Scott Reed}, \bibinfo{person}{Dragomir Anguelov},
  \bibinfo{person}{Dumitru Erhan}, \bibinfo{person}{Vincent Vanhoucke}, {and}
  \bibinfo{person}{Andrew Rabinovich}.} \bibinfo{year}{2015}\natexlab{}.
\newblock \showarticletitle{Going Deeper with Convolutions}. In
  \bibinfo{booktitle}{\emph{Computer Vision and Pattern Recognition (CVPR)}}.
  \bibinfo{pages}{1--9}.
\newblock


\bibitem[\protect\citeauthoryear{Szegedy, Vanhoucke, Ioffe, Shlens, and
  Wojna}{Szegedy et~al\mbox{.}}{2016}]%
        {inception3}
\bibfield{author}{\bibinfo{person}{Christian Szegedy}, \bibinfo{person}{Vincent
  Vanhoucke}, \bibinfo{person}{Sergey Ioffe}, \bibinfo{person}{Jonathon
  Shlens}, {and} \bibinfo{person}{Zbigniew Wojna}.}
  \bibinfo{year}{2016}\natexlab{}.
\newblock \showarticletitle{Rethinking the Inception Architecture for Computer
  Vision}. In \bibinfo{booktitle}{\emph{2016 IEEE Conference on Computer Vision
  and Pattern Recognition (CVPR)}}. \bibinfo{pages}{2818--2826}.
\newblock


\bibitem[\protect\citeauthoryear{Tan, Chen, Pang, Vasudevan, and Le}{Tan
  et~al\mbox{.}}{2019}]%
        {mnasnet}
\bibfield{author}{\bibinfo{person}{Mingxing Tan}, \bibinfo{person}{Bo Chen},
  \bibinfo{person}{Ruoming Pang}, \bibinfo{person}{Vijay Vasudevan}, {and}
  \bibinfo{person}{Quoc~V. Le}.} \bibinfo{year}{2019}\natexlab{}.
\newblock \showarticletitle{MnasNet: Platform-Aware Neural Architecture Search
  for Mobile}. In \bibinfo{booktitle}{\emph{CVPR}}.
\newblock


\bibitem[\protect\citeauthoryear{Truong, Barik, Totoni, Liu, Markley, Fox, and
  Shpeisman}{Truong et~al\mbox{.}}{2016}]%
        {latte}
\bibfield{author}{\bibinfo{person}{Leonard Truong}, \bibinfo{person}{Rajkishore
  Barik}, \bibinfo{person}{Ehsan Totoni}, \bibinfo{person}{Hai Liu},
  \bibinfo{person}{Chick Markley}, \bibinfo{person}{Armando Fox}, {and}
  \bibinfo{person}{Tatiana Shpeisman}.} \bibinfo{year}{2016}\natexlab{}.
\newblock \showarticletitle{Latte: A Language, Compiler, and Runtime for
  Elegant and Efficient Deep Neural Networks}.
\newblock \bibinfo{journal}{\emph{SIGPLAN Not.}}  \bibinfo{volume}{51}
  (\bibinfo{year}{2016}), \bibinfo{pages}{209--223}.
\newblock


\bibitem[\protect\citeauthoryear{Wang, Ananthanarayanan, Zeng, Goel, Pathania,
  and Mitra}{Wang et~al\mbox{.}}{2019}]%
        {wang2019high}
\bibfield{author}{\bibinfo{person}{Siqi Wang}, \bibinfo{person}{Gayathri
  Ananthanarayanan}, \bibinfo{person}{Yifan Zeng}, \bibinfo{person}{Neeraj
  Goel}, \bibinfo{person}{Anuj Pathania}, {and} \bibinfo{person}{Tulika
  Mitra}.} \bibinfo{year}{2019}\natexlab{}.
\newblock \showarticletitle{High-throughput cnn inference on embedded arm big.
  little multi-core processors}.
\newblock \bibinfo{journal}{\emph{IEEE Transactions on Computer-Aided Design of
  Integrated Circuits and Systems}} (\bibinfo{year}{2019}).
\newblock


\bibitem[\protect\citeauthoryear{Wen, Anderson, Radu, O'Boyle, and Gregg}{Wen
  et~al\mbox{.}}{2020}]%
        {wen2020taso}
\bibfield{author}{\bibinfo{person}{Yuan Wen}, \bibinfo{person}{Andrew
  Anderson}, \bibinfo{person}{Valentin Radu}, \bibinfo{person}{Michael~FP
  O'Boyle}, {and} \bibinfo{person}{David Gregg}.}
  \bibinfo{year}{2020}\natexlab{}.
\newblock \showarticletitle{TASO: Time and Space Optimization for
  Memory-Constrained DNN Inference}.
\newblock \bibinfo{journal}{\emph{arXiv preprint arXiv:2005.10709}}
  (\bibinfo{year}{2020}).
\newblock


\bibitem[\protect\citeauthoryear{Wen, Anderson, Radu, O’Boyle, and Gregg}{Wen
  et~al\mbox{.}}{2019}]%
        {wen2019poster}
\bibfield{author}{\bibinfo{person}{Yuan Wen}, \bibinfo{person}{Andrew
  Anderson}, \bibinfo{person}{Valentin Radu}, \bibinfo{person}{Michael~FP
  O’Boyle}, {and} \bibinfo{person}{David Gregg}.}
  \bibinfo{year}{2019}\natexlab{}.
\newblock \showarticletitle{POSTER: Space and Time Optimal DNN Primitive
  Selection with Integer Linear Programming}. In
  \bibinfo{booktitle}{\emph{PACT}}. IEEE.
\newblock


\bibitem[\protect\citeauthoryear{Xie, Girshick, Doll{\'a}r, Tu, and He}{Xie
  et~al\mbox{.}}{2016}]%
        {resnext}
\bibfield{author}{\bibinfo{person}{Saining Xie}, \bibinfo{person}{Ross~B.
  Girshick}, \bibinfo{person}{Piotr Doll{\'a}r}, \bibinfo{person}{Zhuowen Tu},
  {and} \bibinfo{person}{Kaiming He}.} \bibinfo{year}{2016}\natexlab{}.
\newblock \showarticletitle{Aggregated Residual Transformations for Deep Neural
  Networks}.
\newblock \bibinfo{journal}{\emph{2017 IEEE Conference on Computer Vision and
  Pattern Recognition (CVPR)}} (\bibinfo{year}{2016}),
  \bibinfo{pages}{5987--5995}.
\newblock


\bibitem[\protect\citeauthoryear{Zhang, Zhou, Lin, and Sun}{Zhang
  et~al\mbox{.}}{2017}]%
        {shufflenet}
\bibfield{author}{\bibinfo{person}{Xiangyu Zhang}, \bibinfo{person}{Xinyu
  Zhou}, \bibinfo{person}{Mengxiao Lin}, {and} \bibinfo{person}{Jian Sun}.}
  \bibinfo{year}{2017}\natexlab{}.
\newblock \showarticletitle{ShuffleNet: An Extremely Efficient Convolutional
  Neural Network for Mobile Devices}.
\newblock \bibinfo{journal}{\emph{2018 IEEE/CVF Conference on Computer Vision
  and Pattern Recognition}} (\bibinfo{year}{2017}),
  \bibinfo{pages}{6848--6856}.
\newblock


\end{thebibliography}

\clearpage

\appendix
\begin{table*}[t]
	\centering
	\caption{List of convolutional primitives considered.}
	\label{tab:ap_prims}
	\begin{tabular}{@{}lllclll@{}}
		\toprule
		Family & Index & Full Name & \phantom{abc} & Family & Index & Full Name \\
		\midrule
		im2 & a & im2col-copy-self-ab-ki & & direct-sum2d & a & direct-sum2d \\
		& b & im2col-copy-self-atb-ik & & & & \\
		& c & im2col-copy-self-atb-ki & & wino3 & a & winograd-2-3 \\
		& d & im2col-copy-self-atbt-ik & & & b & winograd-2-3-vec-4 \\
		& e & im2col-copy-short-ab-ki & & & c & winograd-2x2-3x3 \\
		& f & im2col-copy-short-atb-ik & & & d & winograd-2x2-3x3-vec-16 \\
		& g & im2col-copy-short-atb-ki & & & e & winograd-2x2-3x3-vec-4 \\
		& h & im2col-copy-short-atbt-ik & & & f & winograd-2x2-3x3-vec-8 \\
		& i & im2col-scan-ab-ki & & & g & winograd-3-3 \\
		& j & im2col-scan-atb-ik & & & h & winograd-3-3-vec4\\
		& k & im2col-scan-atb-ki & & & i & winograd-3x3-3x3 \\
		& l & im2col-scan-atbt-ik & & & j & winograd-3x3-3x3-vec-16 \\
		& m & im2row-copy-short-ab-ik & & & k & winograd-3x3-3x3-vec-4 \\
		& n & im2row-copy-short-abt-ik & & & l & winograd-3x3-3x3-vec-8\\
		& o & im2row-copy-short-abt-ki & & & m & winograd-4x4-3x3 \\
		& p & im2row-copy-short-atbt-ki & & & n & winograd-4x4-3x3-vec-16 \\
		& q & im2row-scan-ab-ik & & & o & winograd-4x4-3x3-vec-4 \\
		& r & im2row-scan-abt-ik & & & p & winograd-4x4-3x3-vec-8 \\
		& s & im2row-scan-abt-ki & & & & \\
		& t & im2row-scan-atbt-ki & & wino5 & a & winograd-2-5 \\
	    & & & & & b & winograd-2-5-vec-4 \\
		kn2 & a & kn2col & & & c & winograd-2x2-5x5 \\
		& b & kn2col-as & & & d & winograd-2x2-5x5-vec-16 \\
		& c & kn2row & & & e & winograd-2x2-5x5-vec-4 \\
		& d & kn2row-aa-ab & & & f & winograd-2x2-5x5-vec-8 \\
		& e & kn2row-aa-abt & & & g & winograd-5-5 \\
		& f & kn2row-aa-atb & & & h & winograd-5-5-vec4 \\
		& g & kn2row-aa-atbt & & & i & winograd-3x3-5x5 \\
		& h & kn2row-as & & & j & winograd-3x3-5x5-vec16 \\
		& & & & & k & winograd-3x3-5x5-vec4 \\
		conv-1x1 & a & conv-1x1-gemm-ab-ik & & & l & winograd-3x3-5x5-vec8 \\
		& b & conv-1x1-gemm-ab-ki & & & m & winograd-4x4-5x5 \\
		& c & conv-1x1-gemm-abt-ik & & & n & winograd-4x4-5x5-vec16 \\
		& d & conv-1x1-gemm-abt-ki  & & & o & winograd-4x4-5x5-vec4 \\
		& e & conv-1x1-gemm-atb-ik  & & & p & winograd-4x4-5x5-vec8 \\
		& f & conv-1x1-gemm-atb-ki & & & & \\
		& g & conv-1x1-gemm-atbt-ik & & mec & a & mec-col \\
		& h & conv-1x1-gemm-atbt-ki & & & b & mec-row-partition \\
		\bottomrule
	\end{tabular}
\end{table*}

\begin{table*}
	\centering
	\caption{List of architectures used to extract common values for convolutional parameters triplets $c$, $k$ and $im$.}
	\label{tab:ap_archs}
	\begin{tabular}{@{}lcl@{}}
		\toprule
		Architecture & \phantom{abc} & Architecture \\
		\midrule
		Alexnet \cite{alexnet} && VGG (11, 13, 16, 19) \cite{vgg} \\
		Inception (v1, v3) \cite{googlenet, inception3} && DenseNet (121, 161, 169, 201) \cite{densenet} \\
		ResNet (18, 34, 50, 101, 152) \cite{resnet} && ResNext (50 32x4d, 101 32x8d) \cite{resnext}\\
		SqueezeNet (1\_0, 1\_1) \cite{squeezenet} && Shufflenet v2 (x0\_5, x1\_0, x1\_5, x2\_0) \cite{shufflenet} \\ 
		MobileNet \cite{mobilenets} && \\
		\bottomrule
	\end{tabular}
\end{table*}

\end{document}